\documentclass{IEEEtaes}

\usepackage[english]{babel}
\usepackage{amsmath,amsfonts,amssymb}
\usepackage{amsthm}
\usepackage{graphicx, epstopdf}
\usepackage[colorinlistoftodos]{todonotes}
\usepackage{subfig}
\usepackage{amsfonts}

\usepackage{url}
\usepackage{siunitx}
\usepackage{algorithm}
\usepackage{algorithmic}
\usepackage[T1]{fontenc}
\usepackage{mathtools}

\usepackage{multirow}
\usepackage{tabularx} % allow set table width
\usepackage{booktabs}
\usepackage{threeparttable}
\usepackage{tablefootnote}

\usepackage{xcolor}
\usepackage{soul}

\pubyear{2023}
\doiinfo{10.1109/TAES.2023.3279353}

\newcommand{\R}{\textbf{R}}

\newcommand{\C}{\textbf{C}}

\newcommand{\bs}{\textbf{s}}
\newcommand{\bc}{\textbf{c}}
\newcommand{\bu}{\textbf{u}}

\newcommand{\Ru}{\R_{\bu}}

\newcommand{\setS}{$\textbf{S}$\:}

\newcommand{\B}{\mathbb{B}}

\newcommand{\Qang}{Q_{\rm{ang}}}
\newcommand{\Qros}{Q_{\rm{ROSIA}}}

\newcommand{\uQang}{\overline{Q}_{\rm{ang}}}
\newcommand{\uQros}{\overline{Q}_{\rm{ROSIA}}}

\newtheorem{lem}{Lemma}{\bfseries}{\itshape}
\usepackage{amsmath}

\begin{document}
\title{ROSIA: Rotation-Search-Based Star Identification Algorithm}
% \author{\IEEEauthorblockN{Chee-Kheng Chng\IEEEauthorrefmark{1}, Chee-Kheng Chng\IEEEauthorrefmark{2}}}
% \thanks{}

\author{Chee-Kheng Chng}
\affil{Australian Institute for Machine Learning, University of Adelaide} 

\author{Álvaro Parra Bustos}
\affil{Australian Institute for Machine Learning, University of Adelaide} 

\author{Benjamin McCarthy}
\affil{University of Adelaide} 

\author{Tat-Jun Chin}
\affil{Australian Institute for Machine Learning, University of Adelaide} 

\receiveddate{Manuscript received 26 September 2022; revised 5 Febrary 2023; accepted 7 May 2023.}

\corresp{\itshape (Corresponding author: Chee-Kheng Chng)}

\authoraddress{Authors' email addresses (same sequence as above)): cheekheng.chng@adelaide.edu.au; alvaro.parra@gmail.com; benmccarthy301@gmail.com; tat-jun.chin@adelaide.edu.au}
% \author{
% Chee-Kheng Chng,
% Alvaro Parra Bustos,
% Ben McCarthy,
% Tat-Jun Chin\\
% % List of institutions
% \and Australian Institute for Machine Learning (AIML), \\
% University of Adelaide
% }

% \begin{document}

% \supplementary{The supplementary materials are available online at \href{http://ieeexplore.ieee.org}{http://ieeexplore.ieee.org}.}
\markboth{CHNG ET AL.}{Rotation-Search-Based Star-ID}
\maketitle

\begin{abstract}
This paper presents a \textit{rotation-search-based} approach for addressing the star identification (Star-ID) problem. The proposed algorithm, ROSIA, is a heuristics-free algorithm that seeks the optimal rotation that maximally aligns the input and catalog stars in their respective coordinates. ROSIA searches the rotation space systematically with the \textit{Branch-and-Bound} (BnB) method. Crucially affecting the runtime feasibility of ROSIA is the upper bound function that prioritizes the search space. In this paper, we make a theoretical contribution by proposing a tight (provable) upper bound function that enables a 400x speed-up compared to an existing formulation. Coupling the bounding function with an efficient evaluation scheme that leverages \textit{stereographic projection} and the \textit{R-tree} data structure, ROSIA achieves feasible operational speed on embedded processors with state-of-the-art performances under different sources of noise. The source of ROSIA is available at \url{https://github.com/ckchng/ROSIA}.
\end{abstract}
\section{Introduction}
Attitude determination plays an integral role in many space missions. A modern star tracker is a popular system for this task due to its insulation against remote hacking. Furthermore, it is lightweight, highly accurate, and power-efficient, which are all desirable properties for operation on a spacecraft \cite{liebe1995star}. A star tracker determines the attitude of spacecraft from an acquired star image. It has two operating modes: star tracking and \textit{Lost-In-Space}. The main difference is that the former assumes prior attitude information, and the latter does not. We address the more challenging Lost-In-Space (LIS) problem in this work. 

\begin{figure} 
    \centering
	\subfloat{\includegraphics[width=0.46\textwidth]{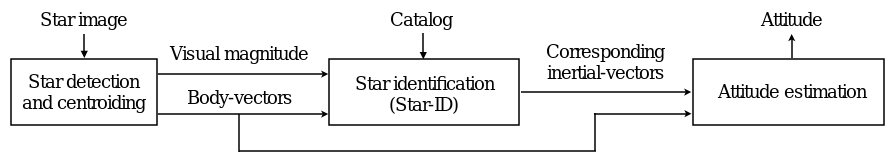}}
	\caption{The established attitude determination pipeline of a star tracker.}\label{fig:star_tracker_pipeline}\vspace{-0.2cm}
\end{figure}

Underpinning a star tracker is a fast and small memory footprint attitude determination algorithm. The standard pipeline for the LIS problem, as illustrated in Fig. \ref{fig:star_tracker_pipeline}, has two main components following the star image pre-processing module: Star-ID and attitude estimation. The latter is a solved problem given the correct correspondences between the detected stars and the catalog stars \cite{markley1988attitude, lerner1978three, markley1999estimate, shuster1993survey}. Meanwhile, Star-ID is a harder problem where most of the research efforts were devoted.

Most of the notable advancements of Star-ID up to the year 2020 were captured in two comprehensive surveys \cite{rijlaarsdam2020survey, spratling2009survey}. The surveys reveal that all existing Star-ID algorithms share two common (main) components - feature extraction and database query. Various robust and feasible star pattern representations were proposed - from simple geometrical patterns such as triangle \cite{liebe1993pattern} and pyramid \cite{mortari2001lost} to complex patterns like the binary star map \cite{padgett1997grid} and the radial and cyclic \cite{wei2019starradialcyclic} features. The common goal of these feature extraction algorithms is to project the detected and catalog stars into a \textit{rotation-invariant space} where the corresponding stars lie on the same \textit{point}. In the application settings where there are measurement uncertainties and outliers (false stars), algorithms were designed to find the \textit{closest match}.

In this work, we explore a new paradigm - searching for the detected-catalog star correspondences directly in the 3D rotation space. Interestingly, the approach was described as a computationally expensive brute-force algorithm in 1977 by Junkins et al.~\cite{junkins1977star}. Another way to interpret their comment is that there were no (known) efficient ways to search directly in the 3D rotation space. Here, we show that such a rotation-search-based Star-ID approach can achieve feasible operational speed on embedded processors with our proposed algorithm, ROSIA.

\begin{figure*}
    \centering
	\subfloat{\includegraphics[width=0.95\textwidth]{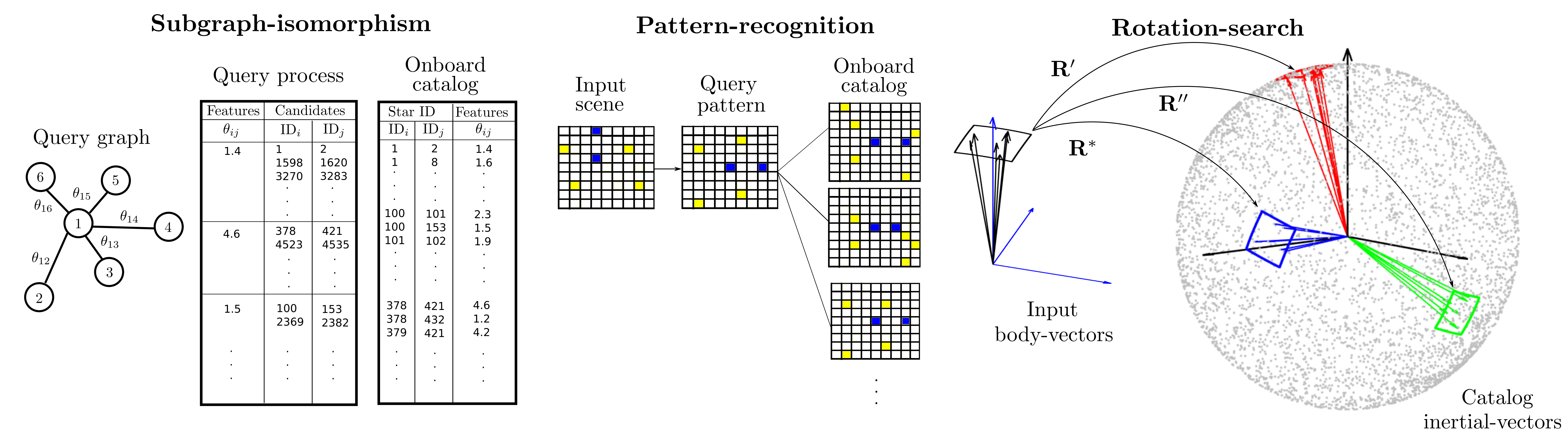}}\\
	\caption{Three different Star-ID paradigms are illustrated. The \textit{subgraph-isomorphism-based} methods (left) query the \textit{onboard catalog} with a subset of its \textit{query graph} (formed with input stars). The feature used is the angular distance $\theta_{ij}$ contained in each edge. The \textit{pattern-recognition-based} methods (middle) first generate a query pattern with the input scene upon pre-processing heuristics, which rely on a subset of the stars (indicated by blue cells). Then, the query pattern is used to query the database, and the closest match is returned. Our proposed \textit{rotation-search-based} approach (right) operates directly on the entire set of input (and catalog) star vectors. It seeks for the optimal rotation that yields the highest number of matches. Each rotation estimate (denoted as $\mathbf{R}'$, $\mathbf{R}''$, and $\mathbf{R}^*$) rotates the input star vectors to query for matches on the unit-sphere that host all the catalog star vectors. The optimal $\mathbf{R}^*$ is associated with the largest intersection set.}\label{fig:paradigms}\vspace{-0.4cm}
\end{figure*}

Such a \textit{direct} method has two strong motivations. Firstly, it solves Star-ID in a top-down fashion, which eliminates the inherent need for voting and verification heuristics in the conventional (bottom-up) \textit{subgraph-isomorphism-based} methods. Secondly, it operates directly on the star vector space, which contains maximum geometrical information. The maximum information regime is also the goal of \textit{pattern-recognition-based} methods, albeit potentially affected by an error-prone pre-processing step. The essence of both the mainstream and our proposed rotation-search-based approaches are illustrated in Fig. \ref{fig:paradigms}, which are compared in Sec. \ref{sec:lit_review}. 
% We compare all three approaches 

The heart of ROSIA is the Branch-and-Bound (BnB) optimization framework~\cite[Chapter 4]{horst2013global}. The BnB framework systematically navigates the rotation space to seek the optimal rotation that maximizes the objective function of ROSIA. Crucially influencing the feasibility of ROSIA is the tightness of its upper bound function, which is used to prioritize \textit{high quality} sub-domains. We took inspiration from the \textit{point cloud registration} community \cite{parra2014fast, hartley2009global} and derived the objective and upper bound functions of ROSIA. Specifically, we propose a novel (provable) upper bound function to address the unique nature of Star-ID - the large mismatch of cardinality between the detected (query) stars and the catalog stars. The derivations and effectiveness of the said objective and upper bound functions are detailed in Sec. \ref{sec:prob_formulation}. 

Having laid down the formulation, Sec. \ref{sec:bnb_rosia} captures two main strategies that we employed to maximize the efficiency of evaluating its objective and upper bound functions. Firstly, ROSIA trims away \textit{infeasible} catalog stars with the geometrical constraint of its objective function and the visual magnitude of stars. Secondly, we cast both evaluations as tree search problems via stereographic projection \cite{needham1998visual} and the R-tree structuring scheme \cite{manolopoulos2006r}.

Empirically, ROSIA demonstrates comparable results to the state-of-the-art Multi-Pole algorithm \cite{schiattarella2017novel} against different sources of noise. The common Star-ID metrics, runtime, and memory consumption were compared and analyzed in Sec. \ref{sec:exp}. 

% \begin{figure} 
%     \centering
% 	\subfloat{\includegraphics[width=8.8cm]{est_star_tracker_pipeline.png}}
% 	\caption{The established attitude determination pipeline of a star tracker.}\label{fig:star_tracker_pipeline}
% \end{figure}

\section{Literature review}\label{sec:lit_review}
Previous Star-ID algorithms can be broadly categorized into two paradigms: subgraph-isomorphism and pattern-recognition approaches.

\subsection{Subgraph-isomorphism-based methods}
Subgraph-based methods pose the Star-ID problem as a graph-matching task. The full graph is constructed with the onboard catalog, where each node is a star, and each edge contains the angular distance between the connecting nodes. The key principle is that the ``query'' graph formed by the input stars is an isomorphic subset of the full graph. To the best of our knowledge, all existing methods in this category seek to match the subgraph with a bottom-up approach. These methods generally perform database queries with the subsets of the input graph \textit{individually} and obtain a consensus result with subsequent voting and verification heuristics. On the contrary, ROSIA utilizes the \textit{complete} query graph\footnote{Each node is connected to all other nodes.} to perform graph matching with a top-down approach. The voting and verification heuristics are manifested in the objective function of ROSIA. We detail the differences between the two approaches below.

One of the earliest matching techniques is to query with a minimal subset of stars. These methods construct primitive geometrical shapes such as triangle \cite{liebe1993pattern, cole2004fast, cole2006fast} and pyramid \cite{mortari2001lost} and extract their geometrical properties, e.g., angular distance and internal angle, as representations. The prerequisite of such methods is to build an onboard catalog with the same geometrical representations. In principle, the onboard catalog has ${M \choose K}$ entries where $M$ is the number of catalog stars, and $K$ is the number of stars in the proposed pattern, e.g., $K=3$ for a triangle. In practical settings, the number of combinations is much smaller when subjected to the restriction on the camera Field of View (FOV). Despite this reduction, the number of potential triangles remains substantial, as seen in the case where $M = 7548$ and FOV = $8^\circ$ with 134,000 potential triangles reported in \cite{padgett1997evaluation}. This highlights a major challenge in these methods - the requirement of a large static memory footprint to store the onboard catalog. In addition, we show in Sec.~\ref{sec:exp} that the memory footprint of these methods scales poorly as the FOV and the number of catalog star increases.    

Similarly, given $N$ stars in an input image, there are ${N \choose K}$ subgraphs to be queried. Inherently, there are two problems: 1) the exponential growth of subgraphs to be queried, and 2) each of the queries returns a large number of candidates due to the lack of distinctness of the primitive shapes. The typical (corresponding) remedies are heuristics such as 1) using only the bright input stars and 2) voting and verification procedures to trim away false candidates. 

To address the problems above, Kosik~\cite{kosik1991star} introduced the \emph{group-match} algorithm. The key idea is to characterize the input stars with a selected \emph{pole} star and its angular distances with respect to all of its neighboring stars. As such, the algorithm seeks to match a group of stars instead of a minimal pattern. Practically, the group-match-based method requires only an onboard catalog with ${M \choose 2}$ entries, each representing a unique pair of stars. Compared to the geometrical-shaped approaches, group-match consumes less static memory since $K=2$. Given the same camera setting above, the database entries are reduced by more than half, with only $66,000$ pairs. The matching is intrinsically more reliable due to the increased information leveraged (all neighboring stars), which tremendously reduces ambiguous matches. Its superiority was confirmed in~\cite{padgett1997evaluation}. 

However, there are three weaknesses in the group match approach. Firstly, the `group matching' is executed in a bottom-up fashion. The group of $N$ stars, defined by a selected pole star and its neighboring stars, is first separated into $N-1$ star-pair. Each pair consists of one of the neighboring stars and the selected pole star. Then, it performs $N-1$ queries with the angular distance of each star-pair. Each query inevitably returns a large number of candidates, which demands voting and verification steps \cite{baldini1993new, van1989star} to identify the correct candidates. Secondly, there is a critical emphasis on selecting the correct pole star, which makes it vulnerable to false stars. Thirdly, the group match approach does not utilize full information of a star pattern - the angular distances between the neighbor stars are not considered. In other words, it is not a complete graph. We show the detrimental effects of such representation in Sec. \ref{sec:exp}. 

To enhance the reliability of the group-match algorithm, Schiattarella et al.~\cite{schiattarella2017novel} introduced the \emph{Multi-Poles} algorithm (MPA) in which multiple detected stars take turns to assume the role of the pole star. Essentially, MPA runs \emph{group-match} (as described above) with multiple pole stars and introduces a series of cross-checking mechanisms to identify correct candidates. Intuitively, MPA is more reliable since it removes the weak link of the group-match algorithm - the heavy reliance on selecting the right pole star. Since the key idea of MPA is running group-match multiple times, it leverages the k-vector technique~\cite{mortari2000k} as its query engine to ensure feasible runtime. Also known as the ``search-less'' algorithm, the k-vector is a catalog \emph{indexing} method that turns angular distance querying into an $\mathcal{O}(1)$ retrieval process.

ROSIA extends the group-match approach in two fundamental aspects. Firstly, it represents the input stars with a complete graph, i.e., leveraging all input stars in their coordinate space, which contains maximum input (geometrical) information. The complete graph is, in principle, an optimal representation of the star pattern with minimum ambiguity. Secondly, ROSIA performs Star-ID with a top-down approach, eliminating conventional voting and verification heuristics. In essence, ROSIA evaluates each rotation estimate with an objective function that leverages the maximum consensus of the input and catalog stars. Practically, ROSIA has the smallest onboard catalog size, with only $M$ entries; each contains the magnitude, position in the inertial coordinate system, and angular distances to two of its closest stars (see Sec.~\ref{sec:prob_formulation}).

\subsection{Pattern-recognition-based methods}
The common aspect of this category is representing star patterns with features other than angular distances. The \emph{grid algorithm} by Padget et al.~\cite{padgett1997grid} is one of the pioneering works in this category. The authors proposed to discretize the input star image into cells and binarize each cell to indicate the presence of stars. The representation (essentially a binary matrix) is then used to query the onboard catalog. The onboard catalog has $M$ entries, where each entry is a binary matrix constructed to represent a star and its neighboring stars (visible within the camera FOV). However, such a representation, on its own, is not rotation invariant. Hence, both the input image and the catalog must be transformed into the same reference frame. The alignment steps are: 1) find a reference star, 2) translate it to the center of the image, 3) find its nearest star, and 4) align the direction vector formed by both stars to the horizontal axis. Later, Na et al.~\cite{na2009modified}, and Aghaei and Moghaddam~\cite{aghaei2016grid} optimized the grid method toward improving robustness.

In the context of Star-ID, pattern-recognition-based features contain the maximum geometrical information available in the input - the entire star pattern is utilized to form the representation. Besides, each star is represented by one pattern; hence it permits a manageable $\mathcal{O}(M)$ database size growth. ROSIA shares both of these strengths. However, the described alignment process is the \emph{Achilles' heel} of pattern-recognition-based methods~\cite{rijlaarsdam2020survey}. One of the major issues with the process is the heavy reliance on selecting a small subset of input stars, which makes it vulnerable to noise and false stars. In contrast, ROSIA does not need an alignment step since it operates directly on the rotation-invariant star vectors' space.

The rest of this section discusses some of the unique patterns which share the above properties. Related in principle to the grid algorithm are the image-based algorithms~\cite{yoon2011new, delabie2013highly}. Yoon et al.~\cite{yoon2011new}'s algorithm can be seen as the smooth version of the grid algorithm. They propose to generate a Gaussian-smoothed synthetic star image based on the pixel coordinates acquired from the centroiding process. Delabie et al.~\cite{delabie2013highly} introduced a shortest-distance-map representation, where each pixel is associated with the distance to the closest star.

Silani and Lovera~\cite{silani2006star} introduced a bar-code feature pattern based on the existence of stars in the discretized area around a chosen pole star given a set of ring strips. Also known as the radial feature, it was further improved by Zhang et al.~\cite{zhang2008full} and Wei et al.~\cite{wei2019starradialcyclic} with the addition of the cyclic feature. These methods rely on lesser pre-processing and require only selecting a reference star to generate a star pattern.

Juang et al.~\cite{juang2004efficient} proposed a singular value representation for star vectors based on the observation that singular values are preserved under rotations. In other words, the body-vectors from the detected stars and the correctly matched inertial-vectors from the catalog stars have identical singular values. This representation also does not require the mentioned alignment process but demands a careful subset selection and sorting based on magnitude information. Again, ROSIA does not need such a pre-processing step. Later, Juang et al.~\cite{juang2012further} revisited their original algorithm to address one of its major limitations - the lack of boresight directions in the onboard catalog. This feature was further pursued recently in~\cite{wei2019star} and \cite{kim2020algorithm}.

\section{Problem Formulation}\label{sec:prob_formulation}
We derive the objective function of ROSIA in this section. The Star-ID task is cast as a rotation search problem in our formulation. The objective function is optimally maximized with the BnB framework, which requires an upper bound function to prioritize and prune the search space \cite{breuel2003implementation}. Both the upper bound function and the search space are detailed below as well. 
 
\subsection{Wahba's problem}
Wahba's problem~\cite{wahba1965least} seeks to find the 3D rotation that relates two reference frames from a set of corresponding vectors. In the context of a star tracker system, the involved frames are the inertial frame of the catalog stars and the body frame of the star tracker. Let $\bs_i$ denote a detected star (unit) vector in the body frame, and $\bc_i$ be its \emph{corresponding} catalog star in the inertial frame. The rotation $\R \in SO(3)$ (Special Orthogonal group for 3D rotation \cite{SO3wiki}) that aligns the body and the catalog frames can be solved by minimizing Wahba's objective

\begin{align}\label{eqn:wahba}
    \sum_{i=1}^{N} \left\| \R \bs_i - \bc_i \right\| ,
\end{align}

\noindent when there are at least two star correspondences, i.e., $N \ge 2$. Solving Wahba's problem is the goal of the attitude estimation module in a star tracker system (Fig. \ref{fig:star_tracker_pipeline}). 

\subsection{Correspondence-free rotation-only point cloud registration}
Correspondence-free rotation-only point cloud registration \cite{parra2014fast, hartley2009global} is the general form of Wahba's problem where the correspondences are unknown. Specifically, the task seeks the 3D rotation that aligns two assumed overlapping point clouds. It is established that correspondence identification and rotation estimation can be solved jointly in a holistic formulation. We state one of the recent formulations \cite{parra2014fast} in the context of Star-ID below, 

\begin{align} \label{eqn:Q_euc}
    Q_{\rm{euc}}(\R) = \sum_{i=1}^{N} {\max_{1 \leq j \leq M}} \lfloor \| \R \bs_i - \bc_j \| \leq \epsilon \rfloor  ,
\end{align}

\noindent where $\epsilon$ is the measurement uncertainty radius, and $\lfloor.\rfloor$ is an indicator function that returns $1$ if the internal relation is true and $0$ otherwise. In words,  $\R \bs_i$ has a match if any of the catalog stars in $\{\bc_j\}^M_{j=1}$ is within its uncertainty vicinity. The max operation is in place to ensure one query star $\textbf{s}_i$ only contributes at most one vote to the overall objective. Since the star vectors are all unit vectors, another equivalent metric is the angular distance, 

\begin{equation}\label{eqn:Q_ang}
    Q_{\rm{ang}}(\textbf{R}) = \sum_{i=1}^{N} {\max_{1 \leq j \leq M}} \lfloor \angle (\textbf{R} \textbf{{s}}_i,\: \textbf{{c}}_j) \leq \alpha_\epsilon \rfloor \: ,
\end{equation}

\noindent where the $\angle(.,.)$ operation yields the angular distance, and $\alpha_\epsilon$ is the angular uncertainty. Associating with the \textit{maximum} $Q_{\rm{ang}}$ is also the largest catalog subset, where the identity of each detected stars can be expressed as $\mathcal{M}^* = \{ \{ i \leftrightarrow j \}^N_{i=1} \: \: \vert \: \: \sum_{i=1}^{N} \max_{1 \leq j \leq M}  \lfloor \angle (\textbf{R}^* \textbf{{s}}_i,\: \textbf{{c}}_j) \leq \alpha_\epsilon \rfloor \: \: \forall \: i \}$. 
% It is instructive to point out that \eqref{eqn:Q_ang} is inherently robust against outliers (false stars in the Star-ID context) since it only matches stars within $\alpha_\epsilon$.

To solve (\ref{eqn:Q_ang}) with BnB, we need an upper bound function to evaluate whether a sub-domain in the rotation space should be further explored (branched). The general idea of BnB is captured in Algorithm \ref{alg:ROSIA}, from line \ref{alg1:line6} to \ref{alg1:line13}. We derive an upper bound function for (\ref{eqn:Q_ang}) in Sec.~\ref{sec:previous_results} after parameterizing the rotation space below.

\begin{algorithm}[t]
	\begin{algorithmic}[1]
		\REQUIRE Scene stars $\{\bs_i, v^{(i)}_s\}_{i=1}^N$, onboard catalog $\{\bc_j,  v^{(j)}_c, \phi^{(j)}_1, \phi^{(j)}_2\}_{j=1}^M$, angular uncertainty $\alpha_\epsilon$, and magnitude uncertainty $\epsilon_{v}$.
		\STATE Extracts triplet features $\{\theta^{(i)}_1, \theta^{(i)}_2\}_{i=1}^N$ from the input scene stars. \label{alg1:line1}
		\STATE Extracts $N$ sub-catalogs $\{\textbf{C}^{(i)}\}_{i=1}^N$. \label{alg1:line2}
		\STATE Stereographically projects and indexes sub-catalogs into $N$ circular R-trees. \label{alg1:line3}
		\STATE Initializes  $q \leftarrow$ empty priority queue, $\B \leftarrow$ cube of side $2 \pi$, $Q^* \leftarrow 0$, $\R^* \leftarrow \emptyset$. \label{alg1:line4}
		\STATE Inserts $\B$ with priority $\uQros(\B)$ into $q$. \label{alg1:line5}
		\WHILE{$q$ is not empty} \label{alg1:line6}
		\STATE Obtain the highest priority cube $\B$ from $q$. \label{alg1:line7}
		\STATE IF $\uQros(\B)$ = $Q^*$, terminate. \label{alg1:line8}
		\STATE  $\Ru \leftarrow$ center rotation of $\B$. \label{alg1:line9}
		\STATE IF $\Qros(\Ru) > Q^*$, $\R^* \leftarrow \Ru$, $Q^* \leftarrow \Qros(\Ru)$. \label{alg1:line10}
		\STATE Subdivides $\B$ into 8 cubes $\{\B_d\}_{d=1}^8$. \label{alg1:line11}
		\STATE For each $\B_d$, IF $\uQros(\B_d) > Q^*$, insert $\B_d$ into $q$ with priority $\uQros(\B_d)$. \label{alg1:line12}
        \ENDWHILE \label{alg1:line13}
        \RETURN $\textbf{M}^*$ and $\R^*$.
	\end{algorithmic}
	\caption{ROSIA : Rotation-search-based Star-ID algorithm}
	\label{alg:ROSIA}
\end{algorithm}

\subsection{Rotation Space Parameterization}\label{sec:rot_space_param}
We parameterize the rotation matrix with the axis-angle representation \cite{axisanglewiki}, where each element in the $SO(3)$ space is mapped to a 3D vector ${\textbf{r}}$. The normalized vector $\hat{{\textbf{r}}}$ represents the axis of rotation, and the magnitude $\parallel {\textbf{r}} \parallel$ represents the rotation angle. The entire $SO(3)$ domain is encapsulated in a $\pi$-ball ($\parallel {\textbf{r}} \parallel$ ranging from 0 to $\pi$) with this parameterization\footnote{Inside the $\pi$-ball, the mapping is one-to-one. On the boundary of the ball, the mapping is two-to-one since ${\textbf{R}_\textbf{r}} = \textbf{R}_{-\textbf{r}}$.}. 

\begin{figure*}
    \centering
	\subfloat{\includegraphics[width=0.92\textwidth]{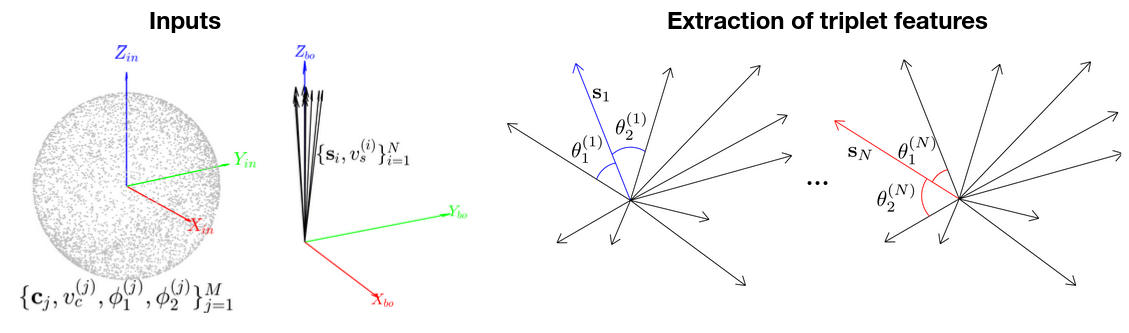}}\\
	\subfloat{\includegraphics[width=0.92\textwidth]{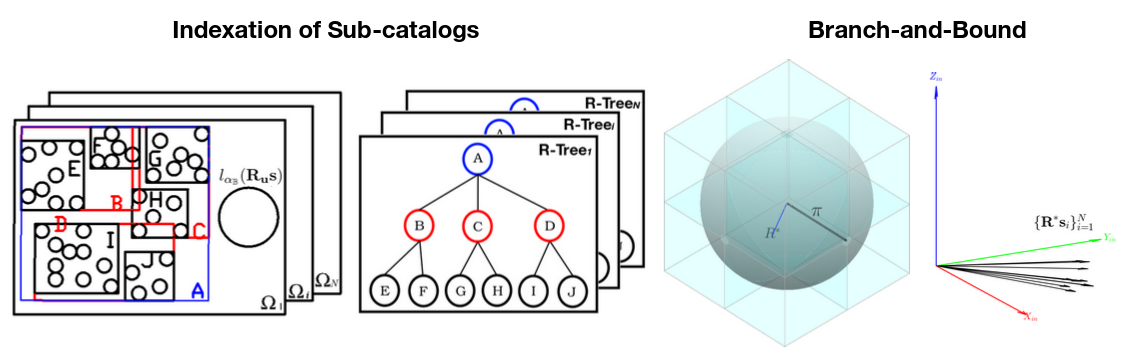}}
	\caption{An overview of ROSIA. The onboard catalog consists of the star vectors $\textbf{c}_j$ in the inertial frame (axes annotated with \textit{in}), visual magnitude $v^{(j)}_c$, and the pre-computed triplet features $\phi^{(j)}_1, \phi^{(j)}_2$. Given a set of body-vectors (axes annotated with \textit{bo}) and visual magnitude of the detected stars $\{s_i, v^{(i)}_s\}^N_{i=1}$, ROSIA first computes the triplet features $\{\theta^{(i)}_1, \theta^{(i)}_2\}^N_{i=1}$. Then, ROSIA retrieves sub-catalogs to be stereographically projected onto $XY$-planes $\{\Omega\}^N_{i=1}$, where each plane is indexed as an R-tree for efficient access. Lastly, ROSIA systematically explores the entire search space (the $\pi$-ball) in a BnB fashion until the optimal rotation $\textbf{R}^*$ is found. The subset of the catalog stars which matches with $\{\textbf{R}^*\textbf{s}_i\}$ informs the ID of each detected star.}\label{fig:bnb_rosia_pipeline}\vspace{-0.4cm}
\end{figure*}

One major component of BnB is the branching of the search space. To facilitate convenient branching, the entire search space, the $\pi$-ball, is bounded in a minimum enclosing cube with the side length of $2\pi$. We denote the cube as $\mathbb{B} = \{{\textbf{u}}, \alpha_\mathbb{B}\}$, where ${\textbf{u}}$ is the center of the cube, and its length to one of the cube vertices is $\alpha_\mathbb{B}$. The branching operation splits each cube evenly into eight sub-cubes (see Algorithm \ref{alg:ROSIA}, line \ref{alg1:line11}). For example, let $\mathbb{B}_0 = \{\textbf{u}_0, \alpha_{\mathbb{B}_0}\}$ denotes the cube that encloses the entire $\pi$-ball. Its center vector ${\textbf{u}_0}$ is the origin, and $\alpha_{\mathbb{B}_0}$ is $\sqrt{3} \pi$. Let $\mathbb{B}'_0$ be one of the (top-right-front) sub-cube from $\mathbb{B}_0$, its center vector $\textbf{u}'_0$ is $(\frac{\sqrt{3}\pi}{2}, \frac{\sqrt{3}\pi}{2}, \frac{\sqrt{3}\pi}{2})$ (different for all sub-cubes) and $\alpha_{\mathbb{B}'_0}$ will be $\frac{\sqrt{3} \pi}{4}$. Fig. \ref{fig:bnb_rosia_pipeline} (second row, third column) illustrates an example of splitting the domain into eight sub-cubes.

\subsection{Previous results}\label{sec:previous_results}
A crucial practical advantage of the axis-angle representation is captured in the inequality below,

\begin{align}\label{eqn:R_ineq}
    \angle(\textbf{R}_{{\textbf{u}}} {\textbf{s}}, \textbf{R}_{{\textbf{r}}} {\textbf{s}}) \leq \parallel {\textbf{u}} - {\textbf{r}} \parallel \:,
\end{align}

\noindent where ${\textbf{s}}$ is a 3D vector, ${\textbf{u}}$ and ${\textbf{r}}$ are two vectors that correspond to the $\R_{{\textbf{u}}}$ and $\R_{{\textbf{r}}}$ rotation matrices. The significance of the inequality above is that the angular distance between two rotated vectors, $\textbf{R}_{{\textbf{u}}} \textbf{s}$ and $\textbf{R}_{{\textbf{r}}} \textbf{s}$, is upper bounded by the Euclidean distance of ${\textbf{u}}$ and ${\textbf{r}}$. The result above was defined in Hartley and Kahl \cite{hartley2009global}.

Given a cube $\mathbb{B}$, it is of interest to determine its maximum angular uncertainty to derive the upper bound of \eqref{eqn:Q_ang}. Leveraging the result in \eqref{eqn:R_ineq} and the fact that the distance of any points in the cube to the center of the cube ${\textbf{u}}$ is shorter than $\alpha_\mathbb{B}$ (by construction), we state the result below,
\begin{equation}\label{eqn:R_ineq_in_sub}
\begin{split}
    \angle(\R_{\bu} \bs, \R_{\textbf{r}} \bs) \; & \leq \; \underset{\textbf{r} \in \B}{\max} \| \bu - \textbf{r} \| \\
                                          & \coloneqq \| \bu - \textbf{v} \| = \alpha_\B \;.
\end{split}
\end{equation}

\noindent where $\textbf{v}$ is one of the vertices of the cube. As such, the upper bound of \eqref{eqn:Q_ang} given $\mathbb{B} = \{{\textbf{u}}, \alpha_\mathbb{B}\}$ is

\begin{equation}\label{eqn:u_Q_ang}
    \overline{Q}_{\rm{ang}}(\mathbb{B}) = \sum_{i=1}^{N} {\max_{1 \leq j \leq M}} \lfloor \angle( \R_{\textbf{u}} \textbf{{s}}_i, \: \textbf{{c}}_j) \leq \alpha_\epsilon + \alpha_\mathbb{B} \rfloor \: .
\end{equation}

\noindent In words, \eqref{eqn:u_Q_ang} bounds the maximum number of \textit{matches} given any rotation matrices in $\mathbb{B}$. Practically, $\mathbb{B}$ with higher $\overline{Q}_{\rm{ang}}$ should be prioritized in the search (best-first-search). Meanwhile, $\mathbb{B}$ with $\overline{Q}_{\rm{ang}}$ lower than the current best estimate can be pruned away safely since it does not contain a better (rotation) solution.

\begin{figure}
    \centering
	\subfloat{\includegraphics[width=0.48\textwidth]{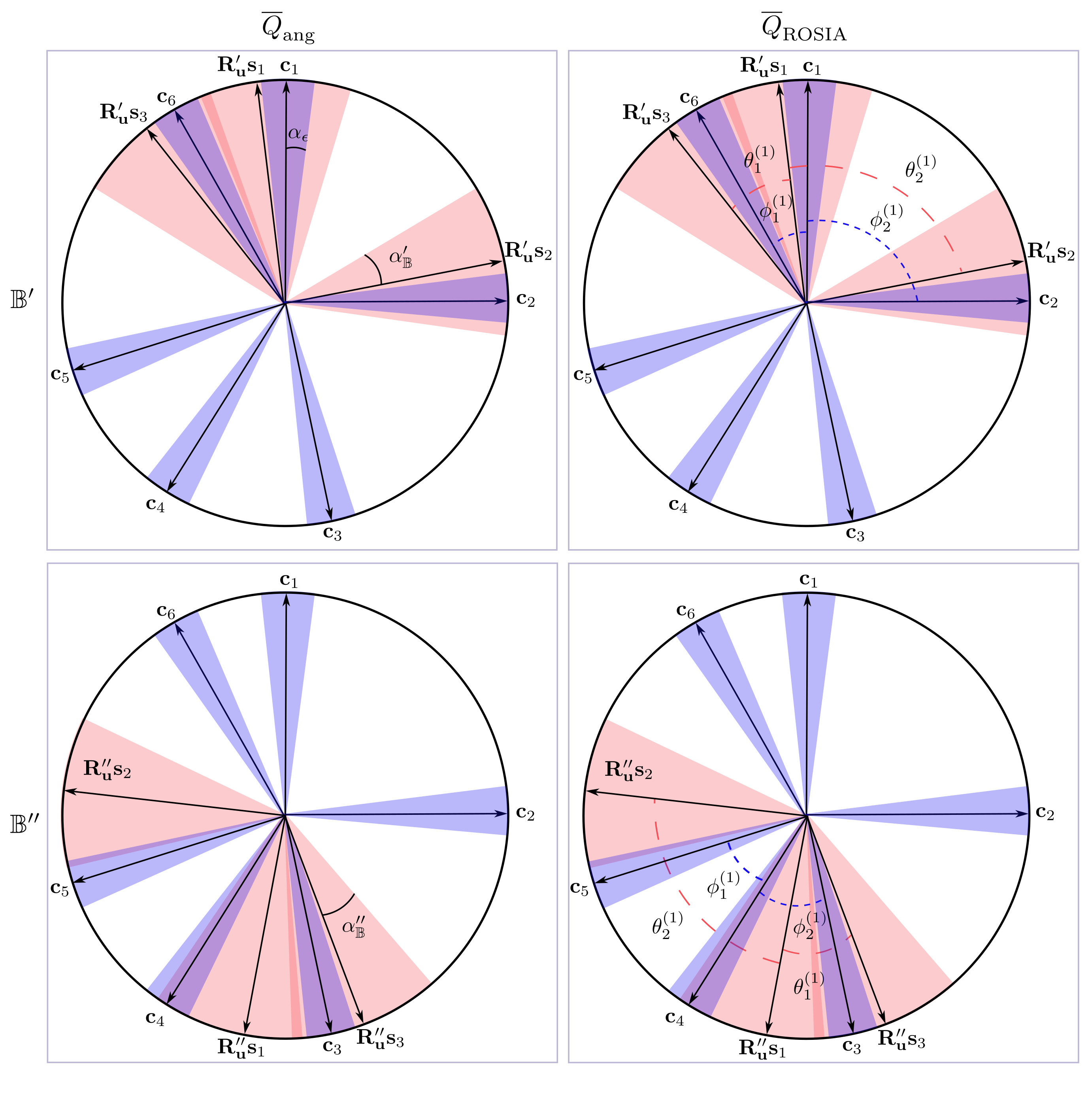}}
	\caption{Comparing the upper bound functions $\uQang$ (left column) and $\uQros$ (right column) on a 2D rotation search example. Given two different sub-domain $\mathbb{B}' = \{\textbf{R}'_\textbf{u}, \alpha'_\B\}$, $\mathbb{B}'' = \{\textbf{R}''_\textbf{u}, \alpha''_\B\}$, the query star vectors $\{\bs_1, \bs_2, \bs_3 \}$ are rotated to two different segments that host different catalog star vectors $\{\bc_1, \ldots, \bc_6 \}$. Pink disk sectors illustrate the angular uncertainty of a sub-domain $\alpha_\B$ and violet sectors the measurement uncertainty of a catalog star vector $\alpha_\epsilon$. In the top row examples, $(\bs_1, \bc_1)$ is a match under $\mathbb{B}'$ (it aligns $\bs_1$ to $\bc_1$ up to $\alpha_\epsilon$ + $\alpha'_\B$ angular distance) for both $\uQang$ and $\uQros$ (where the triplet constraint is fulfilled too). On the contrary, the bottom examples highlight the tightness of $\uQros$ by showing that $(\bs_4, \bc_4)$ is also a match under $\mathbb{B}''$ for $\uQang$, but not for $\uQros$ due to the triplet constraint violation: $\theta^{(1)}_2 \gg \phi^{(1)}_2$; see text for details.}\label{fig:cost_functions}\vspace{-0.5cm}
\end{figure}

subsection{The objective and upper bound functions of ROSIA}\label{subsec:new_obj_upper_bound}
We found that the upper bound in (\ref{eqn:u_Q_ang}) is too conservative for the Star-ID problem. Consequently, a vast majority of the search space is visited, resulting in slow convergence. The crux of the problem is that the query set is a very small subset of the target set, i.e., 21 detected (query) stars on average vs 4934 (evenly distributed) catalog stars. Consequently, it is easy to find a match when each star in the query scene ${\textbf{s}}_i$ is allowed to move within the domain uncertainty $\alpha_\mathbb{B}$ \textit{individually}. 

Such mobility allows the query star pattern to change. Mathematically, the pattern change violates the angular distance preserving property of a rotation. Fig. \ref{fig:cost_functions} (left column) visualizes the described problem. The evaluations of both $\overline{Q}_{\rm{ang}}(\mathbb{B}')$ and $\overline{Q}_{\rm{ang}}(\mathbb{B}'')$ return $3$, i.e., all three query vectors found a match given both sub-domains. Although, it can be clearly seen (in the bottom left) that the geometrical patterns of ${\textbf{c}}_3, {\textbf{c}}_4, {\textbf{c}}_5$ and ${\textbf{s}}_1, {\textbf{s}}_2, {\textbf{s}}_3$ are different. Specifically, the angular distance between ${\textbf{c}}_4, {\textbf{c}}_5$ is significantly smaller than $\angle({\textbf{s}}_1, {\textbf{s}}_2)$.

Based on this key observation, we \textit{anchor} each query star ${\textbf{s}}_i$ with its neighboring stars and formulate the upper bound of ROSIA as

\begin{equation} \label{eqn:u_Q_sid}
\begin{split}
    \uQros(\B) \coloneqq \sum_{i=1}^{N} {\max_{1 \leq j \leq M}} \big(\lfloor &\angle( \R_\bu \bs_i, \: \bc_j) \leq \alpha_\epsilon + \alpha_\B \rfloor  \\
                        &\prod^{K}_{k} \lfloor | \theta_k^{(i)} - \phi_k^{(j)} | \leq 2\alpha_\epsilon \rfloor \big) \: ,
\end{split}
\end{equation}

\noindent where 

\begin{equation}\label{eq:thetas}
    \{\theta_k^{(i)} \: | \: \theta_k^{(i)} \coloneqq \angle(\bs_i, \bs_k), k \neq i \}_{k=1}^{N-1}
\end{equation}
\noindent is the \textit{sorted} angular distances between $i\text{-th}$ query star and its $N-1$ neighboring stars, such that $\theta_k^{(i)} \leq \theta_{k'}^{(i)}$ if $k<k'$, and the equivalent set for the $j\text{-th}$ catalog star is 

\begin{equation}\label{eq:phis}
    \{\phi_k^{(j)} \: | \: \phi_k^{(j)} \coloneqq \angle (\bc_j, \bc_k), k \neq j\}_{k=1}^{M-1}.
\end{equation}

In essence, \eqref{eqn:u_Q_sid} is an extension to \eqref{eqn:u_Q_ang} with neighboring angular distance constraints. The deviation between the sorted neighboring angular distances of the query and catalog stars, $\{|\theta_k^{(i)} - \phi_k^{(j)}|\}^K_{k=1}$, have to be smaller than $2\alpha_\epsilon$ (recall $\alpha_\epsilon$ is the uncertainty of each catalog star) to be considered as a match. Empirically, we found that $K = 2$ offers the best trade-off between speed and identification rate in our simulation settings. Geometrically, it implies \textit{matching a star triplet}; hence we call it the \textit{triplet constraint} henceforth. 

Since BnB requires the objective and bounding functions to have a converging property \cite{horst2013global}, i.e., objective function = upper bound function when the domain collapses to a singleton ($\alpha_\B = 0$), we incorporate the same constraint to \eqref{eqn:Q_ang} and formulate the objective function as 

\begin{equation} \label{eqn:Q_sid}
\begin{split}
    \Qros(\R) \coloneqq \sum_{i=1}^{N} {\max_{1 \leq j \leq M}} &\big(\lfloor \angle (\R \bs_i,\: \bc_j) \leq \alpha_\epsilon \rfloor \\  
                                    &\prod^{K}_{k} \lfloor | \theta_k^{(i)} - \phi_k^{(j)} | \leq 2\alpha_\epsilon \rfloor \big) \: .
\end{split}
\end{equation}

\noindent We provide proof that (\ref{eqn:u_Q_sid}) is a valid upper bound to (\ref{eqn:Q_sid}) in Sections~\ref{Sec:AppendixA} and \ref{Sec:AppendixB}.

The effect of the triple constraint is illustrated in Fig. \ref{fig:cost_functions} (right column). Top right figure depicts a match between ${\textbf{s}}_1$ and ${\textbf{c}}_1$ given
$\mathbb{B}'$, where two nearest stars of ${\textbf{c}}_1$ , ${\textbf{c}}_2$, and ${\textbf{c}}_6$, form a similar pattern with the rotated query stars, $\textbf{R}'_{\textbf{u}}{\textbf{s}}_1$, $\textbf{R}'_{\textbf{u}}{\textbf{s}}_2$, and $\textbf{R}'_{\textbf{u}}{\textbf{s}}_3$. More specifically, $\theta^{(1)}_1$ and $\theta^{(1)}_2$ match with $\phi^{(1)}_1$ and $\phi^{(1)}_2$ (up to $2\alpha_\epsilon$), respectively. The tightness of $\overline{Q}_{\rm{ROSIA}}$ is signified in the bottom right figure, where $\textbf{R}''_{{\textbf{u}}} {\textbf{s}}_1$ and ${\textbf{c}}_4$ are not considered as a match because $\theta^{(1)}_2 >> \phi^{(1)}_2$. Trivially, the added constraining term makes $Q_{\rm{ROSIA}} \leq Q_{\rm{ang}}$ and $\overline{Q}_{\rm{ROSIA}} \leq \overline{Q}_{\rm{ang}}$. In addition, we highlight that since the triplet constraint is not a function of the domain uncertainty angle $\alpha_\mathbb{B}$ in (\ref{eqn:u_Q_sid}), it remains \textit{tight} in the early stage of the search when $\alpha_\mathbb{B}$ is large. As a result, the quality of each subdomain is reflected more precisely with our proposed upper bound $\overline{Q}_\text{ROSIA}$, which leads to much fewer search iterations.

The significance of our proposed formulation is reflected in Tab. \ref{tab:runtime_compare}, where we observed a $\sim400$x speed gain in compared to ${Q}_{\rm{ang}}$ and $\overline{Q}_{\rm{ang}}$. All other algorithmic details are the same apart from the objective and upper bound functions.

\section{ROSIA}\label{sec:bnb_rosia}
Algorithm~\ref{alg:ROSIA} details the steps of ROSIA. In essence, ROSIA branches and bounds the rotation space iteratively until the optimal rotation is found (lines \ref{alg1:line5} - \ref{alg1:line12}). In each iteration, the most computationally expensive operations are the evaluations of $Q_{\rm{ROSIA}}$ (line \ref{alg1:line9}) and $\overline{Q}_{\rm{ROSIA}}$ (line \ref{alg1:line11}). Specifically, given a rotation estimate, evaluating $Q_{\rm{ROSIA}}$ (and $\overline{Q}_{\rm{ROSIA}}$) is to \textit{query} if any of the $M$ catalog stars \textit{matches} (within the uncertainty region) with any of the $N$ rotated query stars. Trivially, a \textit{naive} query implementation takes $\mathcal{O}(NM)$ effort. 

Inspired by \cite{parra2014fast}, we employ two main strategies to maximize query efficiency. Firstly, ROSIA reduces $M$ for each query star (see Sec. \ref{subsec:sub_cat}) by exploiting the geometrical property of triplet constraint and the visual magnitude of the stars. Secondly, we cast the query task as a tree search problem (see Sec. \ref{subsec:tree_search}), which improves the complexity to $\mathcal{O}(N \log M)$ (recall that M $>>$ N).

\begin{table}[!t]
	\begin{center}
		\caption{Comparing average BnB iterations and runtime for ($\Qang$, $\uQang$) against ($\Qros$, $\uQros$).}
		\label{tab:runtime_compare}
		\begin{tabular}{ccc}
			\toprule
            Objective and & Average & Average \\
         upper bound functions & iteration counts & runtime (s)\\
			\noalign{\smallskip}
			\hline
			\noalign{\smallskip}
			$\Qang$, $\uQang$ & $\sim 99000$ & $\sim 6.52$ \\
			$\Qros$, $\uQros$ & $\sim660$ & $\sim0.016$ \\
			\noalign{\smallskip}
			\bottomrule
		\end{tabular}\vspace{-0.2cm}
	\end{center}
\end{table}

\subsection{Strategy 1 - Extraction of Sub-catalogs}\label{subsec:sub_cat}
Here, we describe the key principles and implementation details of extracting $N$ sub-catalogs. In principle, for each query star $\textbf{s}_i$, all $M$ catalog stars are potential matches, each with a different rotation in the $\pi$-ball search space. However, recall the triplet constraint in $Q_{\rm{ROSIA}}$ and $\overline{Q}_{\rm{ROSIA}}$ is not a function of rotation, i.e., the angular distances are rotation-invariant. Therefore, we can enforce the triplet constraints before iterative rotation search begins by discarding \textit{infeasible} matches. Another way to trim $M$ is to leverage the visual magnitude information, which is a common practice for Star-ID algorithms \cite{schiattarella2018efficient}.

In detail, for each query scene, ROSIA first computes and ascendingly sorts all $N \choose 2$ angular distances from the body-vectors $\{\bs_i\}_{i=1}^N$. Then, the first two angular distances of each query star $\{\theta^{(i)}_1, \theta^{(i)}_2$\} (the \textit{triplet feature} henceforth) are extracted, as visualized in Fig. \ref{fig:bnb_rosia_pipeline} (first row, third and fourth columns). The same process is also applied to the catalog stars (extraction of $\{\phi^{(j)}_1, \phi^{(j)}_2$\}) during the onboard catalog construction. Then, for each query star $\textbf{s}_i$, its corresponding sub-catalog is defined as

\begin{equation} \label{eqn:subset}
\begin{split}
    \C_i \coloneqq \{ (\bc_j,\{\phi_k^{(j)}\}_{k=1}^2, \, &{v_{\rm{c}}}^{(j)}) \;|\;  | \phi_k^{(j)} - \theta_k^{(i)} | \leq 2 \alpha_\epsilon \text{ and }\\ 
     &|{v_{\rm{c}}}^{(j)} - {v_{\rm{s}}}^{(i)} | \leq \epsilon_v,\, j=1\ldots M\},
 \end{split}
\end{equation}

\noindent where the first condition is the triplet constraint, and the second condition is the visual magnitude constraint. The visual magnitudes of catalog and query stars are denoted as $v_c$ and $v_s$, respectively, and $\epsilon_v$ is the visual magnitude uncertainty. Let $M_{\rm{sub}}$ be the average size of each sub-catalog, where $M_{\rm{sub}} << M$, the search effort improves to $\mathcal{O}(N M_{\rm{sub}})$. 

Functions $Q_{\rm{ROSIA}}$ and $\overline{Q}_{\rm{ROSIA}}$ given $N$ sub-catalogs $\{\C_i\}^N_{i=1}$ can be expressed as 

\begin{align} \label{eqn:Q_sid_sub}
    Q_{\rm{ROSIA\_sub}}(\R) \coloneqq \sum_{i=1}^{N} {\max_{1 \leq j \leq \vert C_i \vert}} \left(\lfloor \angle (\R \bs_i,\: \bc_j) \leq \alpha_\epsilon \rfloor \right) \: , 
\end{align}

\noindent and 

\begin{align} \label{eqn:u_Q_sid_sub}
    \overline{Q}_{\rm{ROSIA\_sub}}(\B) \coloneqq \sum_{i=1}^{N} {\max_{1 \leq j \leq \vert C_i \vert}} \left(\lfloor \angle( \R_\bu \bs_i, \: \bc_j) \leq \alpha_\epsilon + \alpha_\B \rfloor \right)\: ,
\end{align}

\noindent which are the emphases of \textit{Strategy 2} below.

\subsection{Strategy 2 - R-Tree search}\label{subsec:tree_search}
We elaborate two key ingredients that are needed to cast the $\mathcal{O}(N\:M_{\rm{sub}})$ catalog star query into a $\mathcal{O}(N\: \text{log} \:M_{\rm{sub}})$ tree search problem here - stereographic projection and the R-tree indexation scheme. 

\subsubsection{Stereographic projection to avoid dimension redundancy} We highlight that evaluating $Q_{\rm{ROSIA\_sub}}$ and $\overline{Q}_{\rm{ROSIA\_sub}}$ in the 3D vector space is inefficient. To see this, recall that a star vector is a unit vector that lies on the surface of a unit sphere. Given an angular uncertainty, all possible star positions are contained in a continuous spherical patch, also known as the spherical cap (with circular outline), which is essentially a 2D surface embedded in $\mathbb{S}^2$. 

Inspired by \cite{parra2014fast}, we leverage stereographic projection to map spherical patches onto a 2D plane, which in turn reduces the dimensionality of $Q_{\rm{ROSIA\_sub}}$ and $\overline{Q}_{\rm{ROSIA\_sub}}$ from three to two. We first formally express the spherical patches and the geometrical implication of evaluating $Q_{\rm{ROSIA\_sub}}$ and $\overline{Q}_{\rm{ROSIA\_sub}}$, followed by the essentials of stereographic projection. Interested readers are highly encouraged to read \cite[Chapter 3, Section IV]{needham1998visual} for the full details of stereographic projection. We discuss the essential elements that are crucial to the understanding of our algorithm.

\begin{figure}
    \centering
	\subfloat{\includegraphics[width=0.44\textwidth]{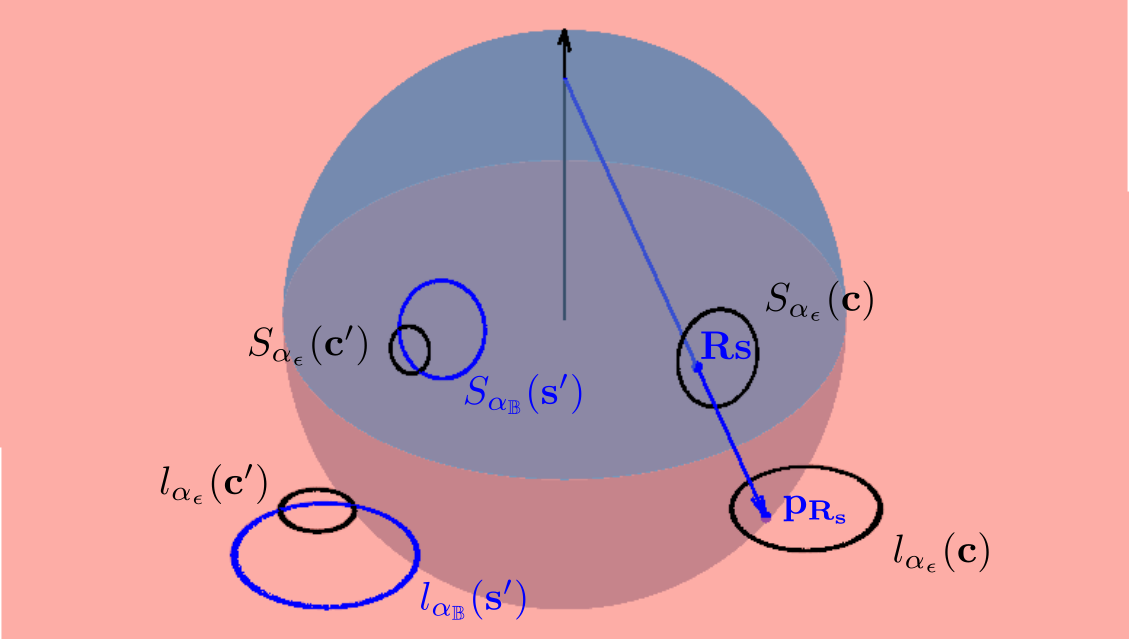}}\\
	\caption{The benefit of stereographic projection in evaluating the objective (and upper bound) function of ROSIA is illustrated. Before projection, evaluating an instance of $Q_{\rm{ROSIA\_sub}}$ is to check if a (rotated) point lies within a spherical patch, which is illustrated by $\textbf{R}\textbf{s}$ and $S_{\alpha_\epsilon}(\textbf{c})$ here. Meanwhile, evaluating $\overline{Q}_{\rm{ROSIA\_sub}}$ is to check if two spherical patches ($S_{\alpha_\epsilon}(\textbf{c}')$ and $S_{\alpha_{\mathbb{B}}}(\textbf{s}')$) intersects. The circular outlines of the spherical patches are highlighted. Upon projecting the spherical patches and point above onto $\Omega$ from $N$, checking if a point lies within a spherical patch is equivalent to checking if the projected point $\textbf{p}_{\textbf{R}\textbf{s}}$ lies in the projected circle $l_{\alpha_\epsilon}(\textbf{c})$; to check if two spherical patches overlap is equivalent to checking if two circles ($l_{\alpha_\epsilon}(\textbf{c}')$ and $l_{\alpha_{\mathbb{B}}}(\textbf{s}')$) intersect. See text for exceptions.}\label{fig:stereo_proj_3D}\vspace{-0.3cm}
\end{figure}

\textbf{Spherical patches.} Formally, a spherical patch defined by a 3D unit vector $\textbf{x}$ and an angular uncertainty $\alpha$ can be expressed as 
\begin{equation}\label{eqn:sph_patch}
    S_{\alpha}(\textbf{x}) = \{\textbf{y} \; | \; \angle(\textbf{y},\textbf{x}) \leq \alpha,\, \| \textbf{y} \| = 1, \| \textbf{x} \| = 1\} \; .
\end{equation}

\noindent There are two types of spherical patches in evaluating $Q_{\rm{ROSIA\_sub}}$ and $\overline{Q}_{\rm{ROSIA\_sub}}$. We denote the patch defined by a catalog star $\textbf{c}_j$ and its measurement angular uncertainty $\alpha_\epsilon$ as $S_{\alpha_\epsilon}(\textbf{c}_j)$, and $S_{\alpha_\mathbb{B}}(\textbf{s}_i)$ represents the patch defined by a query star $\textbf{s}_i$ and the domain angular uncertainty of each cube $\alpha_\mathbb{B}$. Both patches are illustrated in Fig. \ref{fig:stereo_proj_3D}.

\textbf{3D intersection.} Geometrically, evaluating $Q_{\rm{ROSIA\_sub}}$ is equivalent to determining if $\R\bs_i$ (a 3D point) lies within catalog patch $S_{\alpha_\epsilon}(\bc_j)$ (a spherical patch). Fig. \ref{fig:stereo_proj_3D} illustrates an example. Formally, it can be expressed as

\begin{equation}\label{eqn:within}
    \lfloor \R\bs_i \in S_{\alpha_\epsilon}(\bc_j) \rfloor\; . 
\end{equation}

\noindent On the other hand, evaluating $\overline{Q}_{\rm{ROSIA\_sub}}$ is equivalent to determining if
a query patch $S_{\alpha_\B}(\Ru \bs_i)$ intersects with a catalog patch $S_{\alpha_\epsilon}(\bc_j)$, as depicted in Fig. \ref{fig:stereo_proj_3D}. 

\begin{equation}\label{eqn:intersects}
    \lfloor S_{\alpha_\B}(\Ru \bs_i) \cap S_{\alpha_\epsilon}(\bc_j) \neq \emptyset \rfloor\; .
\end{equation}

\textbf{Stereographic projection is conformal}, which preserves circles and circle intersections. Therefore, the circular outlines of spherical patches are projected as circles, and the intersection tasks above (\eqref{eqn:within} and \eqref{eqn:intersects}) can be cast into 2D intersection tasks, which can be solved efficiently. The projection is defined everywhere on the unit sphere apart from the projection point. We describe below all three possible projection outcomes of spherical patches on the 2D projection plane: interior circles, exterior circles, and half-planes.

\textbf{3D point to 2D point.} We first detail the projection of a 3D unit vector (e.g., a query star) onto the $XY\text{-plane}$ $\Omega$. Let $[\varphi \in [0, \pi], \theta \in [0, 2\pi]]$ represent the spherical coordinates of a point on the unit sphere, its stereographic projection onto $\Omega$ with $N = [0, 0, 1]$ as the projection point can be expressed as

\begin{figure}
    \centering
	\subfloat{\includegraphics[width=0.42\textwidth]{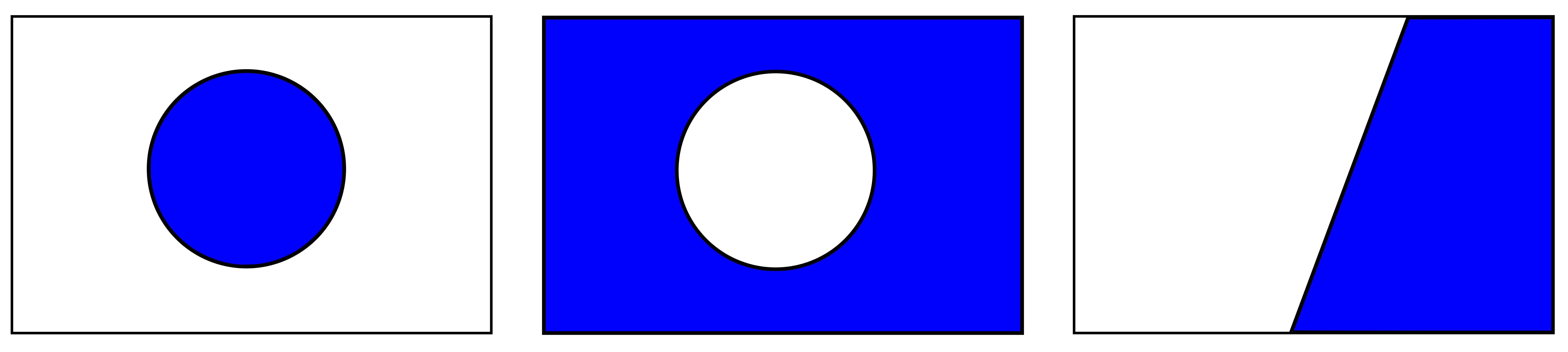}}\\
	\caption{Three possible stereographic projection outcomes of spherical patches. Left: The surface is projected into the \textit{interior of a circle}. Middle: The surface is projected to the \textit{exterior of a circle} when the patch contains the projection point. Right: The surface is projected to \textit{one side of the half-plane} when the projection point touches the boundary of the spherical patch.}\label{fig:sphe_proj_outcomes}\vspace{-0.2cm}
\end{figure}

\begin{figure*}[!t]
    \centering
	\subfloat{\includegraphics[width=0.9\textwidth]{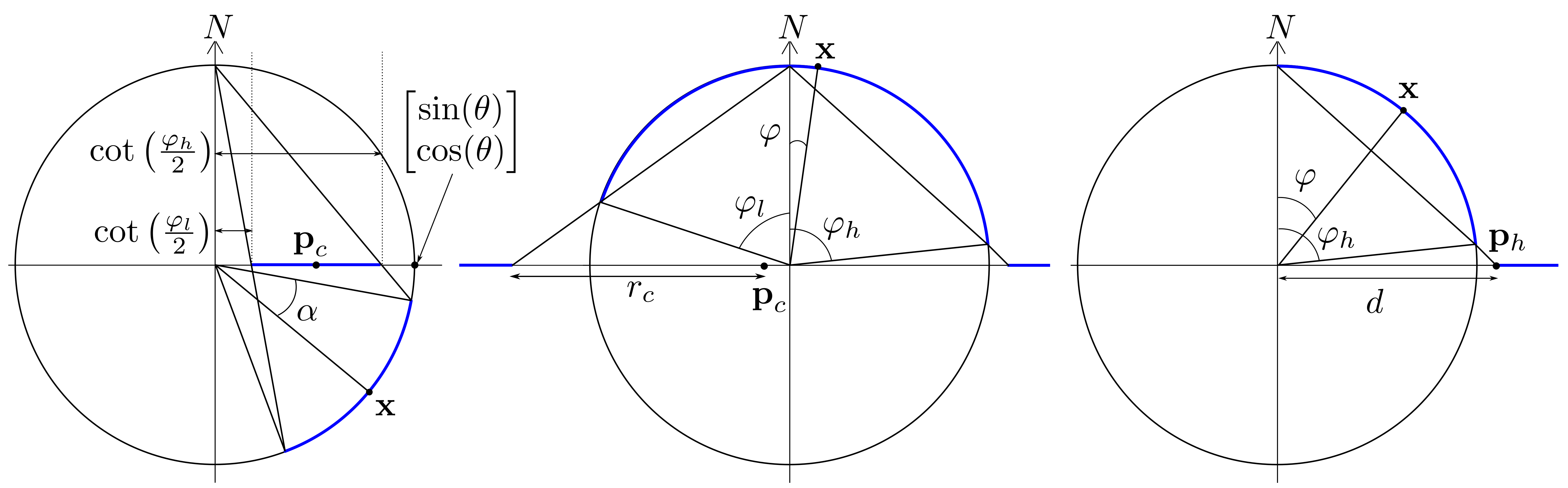}}\\
	\caption{The geometry of all three different projections corresponding to Fig. \ref{fig:sphe_proj_outcomes} in a 2D cross-section view that passes through the north pole $N$, and the center of the spherical patch $\textbf{x}$. The spherical patches (defined by $\textbf{x}$ and $\alpha$) are depicted as blue (bold) arcs, and their corresponding patches on $\Omega$ are highlighted with blue (bold) line segments. The diagram on the left (interior circle projection) relates to \eqref{eqn:center_circle} and \eqref{eqn:radius_circle} by showing that 1) the center of the circle $\textbf{p}_c$ is formed with a direction defined by $\theta$, and 2) the middle point between the radial distances defined by $\varphi_l$ and $\varphi_h$ (which also define the circle radius $r_c$). The exterior circle and half-plane projections are visualized in the middle and right diagrams, respectively. For the half-plane projection, $d$ is the distance to the edge of the half-plane in the direction of $\textbf{p}_h$.}\label{fig:sphe_proj_geo}\vspace{-0.4cm}
\end{figure*}

\begin{equation}\label{eqn:stereo_proj_spherical_coord}
    {\textbf{p}} = \cot \bigg(\frac{\varphi}{2}\bigg) \begin{bmatrix}
                                                    \sin(\theta)\\
                                                    \cos(\theta)
                                                    \end{bmatrix}\: .
\end{equation}

\textbf{Spherical patch to interior circle patch.} When the projection point is not part of the spherical patch, the surface is projected to the interior of a circle patch, as visualized in Fig. \ref{fig:sphe_proj_outcomes} (left). Consistent with the notation above, let $[\varphi, \theta]$ be the star coordinates and $\alpha$ be its angular uncertainty; the 2D center point of the projected circle on $\Omega$ is defined as 

\begin{equation}\label{eqn:center_circle}
    {\textbf{p}}_c = \frac{\cot (\frac{\varphi_h}{2}) + \cot (\frac{\varphi_l}{2})}{2}  \begin{bmatrix}
                                                                                                \sin(\theta)\\
                                                                                                \cos(\theta)
                                                                                                \end{bmatrix} \: ,
\end{equation}

\noindent where $\varphi_h \coloneqq \varphi + \alpha$ and $\varphi_l \coloneqq \varphi - \alpha$ are the upper and lower bounds of the inclination angles, respectively. Meanwhile, the radius of the circle is 

\begin{equation}\label{eqn:radius_circle}
    r_c = \frac{\vert \cot (\frac{\varphi_h}{2}) - \cot (\frac{\varphi_l}{2}) \vert}{2} \: .
\end{equation}

\noindent To see this, recall that stereographic projection preserves circles. Hence, any two points on the opposite end (with maximum angular distance within the patch, i.e., $2 \alpha$) of the spherical patch will be projected to the two (minimal) points on the plane that defines a circular patch. We denote the projected circle as $l_{\alpha}(\textbf{x}) \coloneqq \{\textbf{p}_c, \textbf{r}_c\}$. Fig. \ref{fig:sphe_proj_geo} (left) illustrates the relevant geometry of the projection in a 2D cross-section view.

% \begin{table}
% 	\begin{center}
% 		\caption{Possible patch-patch intersection queries on $\Omega$.}
% 		\label{tab:patch_patch_query}
% 		\begin{tabular}{cc}
% 			\toprule
%             Patch-Patch-intersection & Query equations \\
% 			\noalign{\smallskip}
% 			\hline
% 			\noalign{\smallskip}
%             interior circle and interior circle & $ \lfloor \| \textbf{p}'_c - \textbf{p}_c \| \leq r'_c + r_c \rfloor \: $ \\
% 			 interior circle and exterior circle & $ \lfloor \| \textbf{p}'_c - \textbf{p}_c \| \geq r'_c + r_c \rfloor \: $ \\
% 			  interior circle and half-plane &  $\lfloor {\hat{\textbf{p}}_h}^\intercal \textbf{p}'_c - (d + r'_c)  \geq 0 \rfloor \:\:\: \text{if} \:\:\: \varphi_h < \pi$ \\
% 			   & $\lfloor {\hat{\textbf{p}}_h}^\intercal \textbf{p}'_c - (d + r'_c) < 0 \rfloor \:\:\: \text{if} \:\:\: \varphi_h \geq \pi$ \\
% 			\noalign{\smallskip}
% 			\bottomrule
% 		\end{tabular}\vspace{-0.5cm}
% 	\end{center}
% \end{table}

\textbf{Spherical patch to exterior circle patch.} When the projection point is contained within the spherical patch, the surface is projected to the exterior of a circle patch, as visualized in Fig. \ref{fig:sphe_proj_outcomes} (middle). The circle is also defined by \eqref{eqn:center_circle} and \eqref{eqn:radius_circle}, as we can see in Fig. \ref{fig:sphe_proj_geo} (middle).

\textbf{Spherical patch to half-plane.} When the projection point touches the edge of the spherical patch, i.e., $\varphi_l = 0$, the surface is projected to a half-plane (Fig. \ref{fig:sphe_proj_outcomes} (right)) since the projection of the projection point is at infinity (see \eqref{eqn:stereo_proj_spherical_coord}). The half-plane is defined as 

\begin{equation}
\begin{split}\label{eqn:half_plane}  
    {\hat{\textbf{p}}_h}^\intercal \textbf{p} - d & \geq 0 \:\:\: \text{if} \:\:\: \varphi_h < \pi \\
    {\hat{\textbf{p}}_h}^\intercal \textbf{p} - d & < 0 \:\:\: \text{if} \:\:\: \varphi_h \geq \pi
\end{split}
\end{equation}

\noindent where $\textbf{p}$ is any arbitrary point on $\Omega$, $\hat{\textbf{p}}_h$ is the direction of ${\textbf{p}}_h$, i.e., the projection of the furthest point on the patch from the north pole. The distance from the origin to the edge of the projected half-plane (in direction $\hat{\textbf{p}}_h$) is denoted as $d \coloneqq \cot \big( \frac{\varphi_h}{2} \big)$, and the side of the plane is decided based on $\varphi_h$, which can be seen in Fig. \ref{fig:sphe_proj_geo} (right).

\textbf{2D intersection.} Owing to the intersection preservation property, evaluating an instance of $Q_{\rm{ROSIA\_sub}}$ is equivalent to checking if the (projected) query star (point) lies within a catalog star patch upon projection. Tab. \ref{tab:point_patch_query} tabulates all three different evaluations depending on the projection types of the catalog star. We denote the projected query star as $\textbf{p}'$, and the rest of the projected catalog patch notations are consistent with \eqref{eqn:center_circle}, \eqref{eqn:radius_circle}, and \eqref{eqn:half_plane}.

On the other hand, evaluating an instance of $\overline{Q}_{\rm{ROSIA\_sub}}$ is equivalent to checking if the query star patch intersects with a catalog star patch. Tab. \ref{tab:patch_patch_query} tabulates three necessary combinations to be checked. The notations of the two patches involved are distinguished with $(')$. Any combination that does not include an interior circle is automatically a match since they share (at least) the projection point.\\

\begin{table}[!t]
	\begin{center}
		\caption{Possible point-patch intersection queries on $\Omega$.}
		\label{tab:point_patch_query}
		\begin{tabular}{cc}
			\toprule
            Point-Patch-intersection & Query equations \\
			\noalign{\smallskip}
			\hline
			\noalign{\smallskip}
            point and interior circle & $ \lfloor \| \textbf{p}' - \textbf{p}_c \| \leq r_c \rfloor \: $ \\
			point and exterior circle & $ \lfloor \| \textbf{p}' - \textbf{p}_c \| \geq r_c \rfloor \: $ \\
			point and half-plane &  $\lfloor {\hat{\textbf{p}}_h}^\intercal  \textbf{p}' - d  \geq 0 \rfloor \:\:\: \text{if} \:\:\: \varphi_h < \pi$ \\
			  & $\lfloor {\hat{\textbf{p}}_h}^\intercal \textbf{p}' - d  < 0 \rfloor \:\:\: \text{if} \:\:\: \varphi_h \geq \pi$ \\
			\noalign{\smallskip}
			\bottomrule
		\end{tabular}\vspace{-0.5cm}
	\end{center}
\end{table}

\begin{table}[!t]
	\begin{center}
		\caption{Possible patch-patch intersection queries on $\Omega$.}
		\label{tab:patch_patch_query}
		\begin{tabular}{cc}
			\toprule
            Patch-Patch-intersection & Query equations \\
			\noalign{\smallskip}
			\hline
			\noalign{\smallskip}
            interior circle and interior circle & $ \lfloor \| \textbf{p}'_c - \textbf{p}_c \| \leq r'_c + r_c \rfloor \: $ \\
			 interior circle and exterior circle & $ \lfloor \| \textbf{p}'_c - \textbf{p}_c \| \geq r'_c + r_c \rfloor \: $ \\
			  interior circle and half-plane &  $\lfloor {\hat{\textbf{p}}_h}^\intercal \textbf{p}'_c - (d + r'_c)  \geq 0 \rfloor \: ,  $\\
                                            & $\text{if} \:\:\: \varphi_h < \pi \: ; $\\
			                                 & $\lfloor {\hat{\textbf{p}}_h}^\intercal \textbf{p}'_c - (d + r'_c) < 0 \rfloor \: , $\\
                                            & $\text{if} \:\:\: \varphi_h \geq \pi$ \\
			\noalign{\smallskip}
			\bottomrule
		\end{tabular}\vspace{-0.5cm}
	\end{center}
\end{table}

\subsubsection{Indexation for efficient access} There is a major benefit in indexing the interior circles since they constitute the vast majority of the catalog patches. Exterior circles and half-planes are rare; hence, we store them in a (small) list for scanning as described in \cite{parra2014fast}. We index the interior circles hierarchically in a R-tree structure \cite{manolopoulos2006r}, which cast the evaluations of $Q_{\rm{ROSIA\_sub}}$ and $\overline{Q}_{\rm{ROSIA\_sub}}$ to $\mathcal{O}(N\: \text{log}\: M_{\rm{sub}})$ tree search problems. Then, we leverage the geometrical property of our search space $\pi$-ball to implement \textit{matchlist} \cite{breuel2002comparison}, which reduces the stars to be queried from $N$ to $N_{\rm{avg}}$.

\textbf{Circular R-tree.} Considering only the interior circles of catalog stars, evaluating $Q_{\rm{ROSIA\_sub}}$ and $\overline{Q}_{\rm{ROSIA\_sub}}$ translate to solving point-circle and patch-circle intersections, respectively. The interior circles can be structured geometrically in hierarchical order as visualized in Fig.~\ref{fig:bnb_rosia_pipeline} (second row, first and second columns). From a top-down perspective, each node of the R-tree encodes a Minimum Bounding Rectangle (MBR) that encodes the MBRs represented by its \textit{child} nodes. At the lowest level of each branch, i.e., the \textit{leaf} nodes, each node contains close-by circles that define the MBR. As such, at each tree level apart from the lowest level, checking if a query point (patch) intersects with one of the MBRs is essentially a point-rectangle (patch-rectangle) intersection problem. Efficient evaluation results from ignoring the branches whose MBR does not overlap with the query point or patch. Fig.~\ref{fig:bnb_rosia_pipeline} (third column, top) illustrates an example where the query patch $l_{\alpha_\B}(\R_{\bu} \bs)$ does not intersect with the root of the tree, resulting in immediate query termination. 

\textbf{Matchlists.} The search effort can be further improved by reducing $N$. Given any cube $\mathbb{B}$, the \textit{intersection set} between the query scene stars \setS $\coloneqq \{\textbf{s}_i\}^N_{i=1}$ and their corresponding sub-catalog $\textbf{C}^{(i)}$ is

\begin{equation}\label{eqn:matchnlist}
    I = \{ {\textbf{s}} \in \textbf{S} \:\:\vert\:\: \exists\: {\textbf{c}} \in \textbf{C}^{(i)},  \lfloor \angle( \R_\bu \bs, \: \bc) \leq \alpha_\epsilon + \alpha_\B \rfloor\}  \: .
\end{equation}

\noindent The set $I$ is coined as the \textit{matchlist} of $\mathbb{B}$ by Breuel \cite{breuel2002comparison}. By construction, the matchlist of a sub-cube $\mathbb{B}^{'} \subset \mathbb{B}$, is always a subset of $I$, i.e., $I^{'} \subset I$. To see this, recall that the upper bound function in \eqref{eqn:u_Q_sid} always returns the largest possible matching subset of $\textbf{S}$ within $\mathbb{B}$. Hence, by maintaining a matchlist with each cube, query stars that are not in $I$ can be skipped in the evaluations of $Q_{\rm{ROSIA\_sub}}(\R_{u}^{'})$ and $\overline{Q}_{\rm{ROSIA\_sub}}(\mathbb{B}^{'})$.  Since the size of the matchlist decreases monotonically as ROSIA branches the cube, the search effort reduces alongside. Let $N_{\rm{avg}}$ represents the average size of the matchlists, the evaluations of $Q_{\rm{ROSIA\_sub}}$ (and $\overline{Q}_{\rm{ROSIA\_sub}}$) now takes $\mathcal{O}(N_{\rm{avg}}\: \text{log}\: M_{\rm{sub}})$ effort.

\subsection{Computational complexity}
Tab.~\ref{tab:comp_analysis} summarizes the computational cost of the main components in ROSIA. There are two main steps in the triplet feature extraction process: 1) the computation of ${N \choose 2}$ angular distances, and 2) sorting $N$ angular distances for $N$ query stars which takes $\mathcal{O}({N} \log {N})$ effort. For building circular R-tree, it involves $\mathcal{O}({N \log M})$ complexity to first retrieve $N$ subsets from a magnitude-sorted catalog of size $M$, as defined in~\eqref{eqn:subset}. Then, for each sub-catalog of cardinality $M_{\rm{sub}}$, stereographic projections and tree insertion can be done in linear time ($\mathcal{O} (M_{\rm{sub}}))$. Note that all the above operations are executed only once before the BnB search begins. At each iteration of BnB, the evaluation of $Q_{\rm{ROSIA\_sub}}$ and $\overline{Q}_{\rm{ROSIA\_sub}}$ take $\mathcal{O}(N_{\rm{avg}}  \log  M_{\rm{sub}})$ time each.

\subsection{Space complexity}\label{subsec:space_comp}
The onboard catalog has $M$ entries, and each entry has six values: three for the inertial-vector coordinates $\bc$, two for the triplet feature  $(\phi_1, \phi_2)$, and one for the visual magnitude $v_{\rm{c}}$. Hence, the fixed storage space has $\mathcal{O}(M)$ complexity. The maintenance of $N$ sub-catalogs is performed with pointers to avoid memory redundancy.

\begin{table}
	\begin{center}
		\caption{Computational complexity of ROSIA.}
		\label{tab:comp_analysis}
		\begin{tabular}{cc}
			\toprule
            Component & Complexity\\
			\noalign{\smallskip}
			\hline
			\noalign{\smallskip}
			Extraction of Triplet features & $\mathcal{O}(N^2 + N \log N)$\\
		    Extraction of Sub-catalogs & $\mathcal{O}(N  \log  M)$\\ 
			Stereographic projection & $\mathcal{O}(N M_{\rm{sub}})$\\
			Circular R-tree building & $\mathcal{O}(N M_{\rm{sub}})$\\
			$Q_{\rm{ROSIA\_sub}}$ evaluation & $\mathcal{O}(N_{\rm{avg}} \, \text{log} \,  M_{\rm{sub}})$\\
			$\overline{Q}_{\rm{ROSIA\_sub}}$ evaluation & $\mathcal{O}(N_{\rm{avg}} \, \text{log} \,  M_{\rm{sub}})$ \\
			\noalign{\smallskip}
			\bottomrule
		\end{tabular}\vspace{-0.5cm}
	\end{center}
\end{table}

\section{Experiments} \label{sec:exp}
We evaluated ROSIA with both simulated and real data in this section. We first conducted controlled experiments using simulated data to analyze the performance of ROSIA against different noise sources thoroughly. In addition, we compared ROSIA with the state-of-the-art Multi-Poles algorithm (MPA) \cite{schiattarella2017novel}. To enhance the performance of MPA, we incorporated the visual magnitude information, as described in \cite{schiattarella2018efficient}. We report the identification rate, the rate of no-result, and the false positive rate for both algorithms.

In addition, we also report the average runtime and memory consumption of ROSIA to showcase its feasibility. ROSIA was implemented in C++. We report the runtime results on both x86 and ARM CPU architectures. The runtime results were recorded on the following commercialized-off-the-shelf (COTS) devices:

\begin{itemize}
    \item a desktop powered by an Intel i5-8400 2.8 GHz x86 64-bit processor (i5 henceforth),
    \item Rapsberry Pi 4 Model B, which contains a 64-bit Quad-core Cortex-A72 ARM v7 CPU with a clock speed of 1.5GHz (A72 henceforth), and
    \item Nvidia Jetson AGX Xavier, which contains a 64-bit 8-core Nvidia Carmel ARM v8.2 CPU with a clock speed of 2.2GHz (Carmel henceforth).
\end{itemize}

The result of i5 was provided to ease future benchmarking and the latter two were included mainly to showcase the practicality of ROSIA on different ARM processor tiers that are employed in CubeSat-qualified devices\footnote{See \url{https://kplabs.space/leopard} and \url{https://satsearch.co/products/innoflight-cfc-500-tflop-flight-computer-payload-processor}.} today. Furthermore, we highlight that ROSIA is implemented to run on a single core without the concurrency mechanism. We refer to these devices with their processor names in the result discussions.

\subsection{Simulated data experiments}
\textbf{Simulation setup.} We generated our input data with the scripts released by the organizers of the European Space Agency (ESA) \emph{Star Trackers: First contact} competition\footnote{https://kelvins.esa.int/star-trackers-first-contact/scripts/}. The HIPPARCOS catalog \cite{perryman1997hipparcos} is utilized in the data generation. The camera in the simulation has the following configurations: $14^\circ \times 14^\circ$ FOV, $1024 \times 1024$ pixel resolutions, and a visual magnitude sensitivity of 6. Under this setting, there are 21 stars on average in each image and 4934 catalog stars after binary stars removal. We evaluated different aspects of ROSIA, namely its robustness against positional noise, visual magnitude noise, and false stars. We generated 1000 images with random orientations for each noise level. 

\textbf{Hyperparameters.} ROSIA and MPA share two hyperparameters: the angular distance deviation threshold\footnote{For MPA, 2$\alpha_\epsilon$ is used to query the catalog, similar to our triplet constraint.} $\alpha_\epsilon$ and the magnitude deviation threshold\footnote{Note that $\alpha_\epsilon$ and $\epsilon_v$ were referred to as angular and visual magnitude uncertainties prior to this. Calling them `thresholds' is more intuitive from a hyperparameters perspective.} $\epsilon_v$. On top of that, MPA has two more hyperparameters that serve its verification heuristics: 1) the minimum number of verified stars $th^*$, which allows the algorithm to proceed to the confirmation stage if it is exceeded, and 2) the maximum number of chosen poles $R_p$, where the algorithm halts and returns \textit{no-result} if it is exceeded. Based on the recommendation in~\cite{schiattarella2018efficient}, we set $th^*=3$. We set $R_p=20$, which is larger than the minimum recommended of 6~\cite{schiattarella2017novel} because our simulation is harsher with a smaller FOV. 

We evaluated both algorithms on three sets of (common) hyperparameters: 1) $\mathcal{S}_1 = \{\alpha_\epsilon = \ang{0.0205}, \epsilon_v = 0.45\}$, 2) $\mathcal{S}_2 = \{\alpha_\epsilon = \ang{0.0275} , \epsilon_v = 0.6\}$ and 3) $\mathcal{S}_3 = \{\alpha_\epsilon = \ang{0.0275} , \epsilon_v = 1.2\}$. We highlight that the visual magnitude threshold function proposed in~\cite{schiattarella2018efficient} is not adopted since our experiments involve extreme visual magnitude noise levels, which the authors did not consider. The first two sets, $\mathcal{S}_1$ and $\mathcal{S}_2$, are decided based on the $1.5\sigma$ and $2\sigma$ of the typical noise settings,  i.e., zero-mean and standard deviations (SD) of 1 pixel (or $\ang{0.0136}$) and 0.3 visual magnitudes. We included $\mathcal{S}_3$ to cover both the maximum SD ranges in the positional and visual magnitude noise experiments below. \\

\textbf{Metrics.} The metrics are detailed here. ID rate is the usual metric for star ID, which is the ratio of the successfully recognized image over the total number of images. We follow the definition of success in \cite{schiattarella2018efficient}, i.e., at least three identified stars with no false positive (wrongly identified stars). ROSIA returns a no-result if 1) it fails to match at least three of the query stars or 2) the number of matches is less than $30\%$ of the query stars. Meanwhile, MPA returns a no-result if $R_p$ is exceeded. A result that is neither a success nor a no-result is considered a false positive. \\

\textbf{Positional noise.} We first evaluate the robustness of ROSIA and MPA against positional noise. In this experiment, the detected body-vectors were perturbed with Gaussian noise. We sweep the noise SD from 0 to $0.027^\circ$ (with zero-mean), which is equivalent to 0 to 2 pixels in our simulation setup. Meanwhile, the visual magnitude noise SD is fixed at 0.3. The \textit{missing stars} are the byproducts of both positional (out of FOV) and visual magnitude noise (out of detection threshold).

\begin{figure*}[!t]
    \centering
    \subfloat{\includegraphics[width=0.32\textwidth]{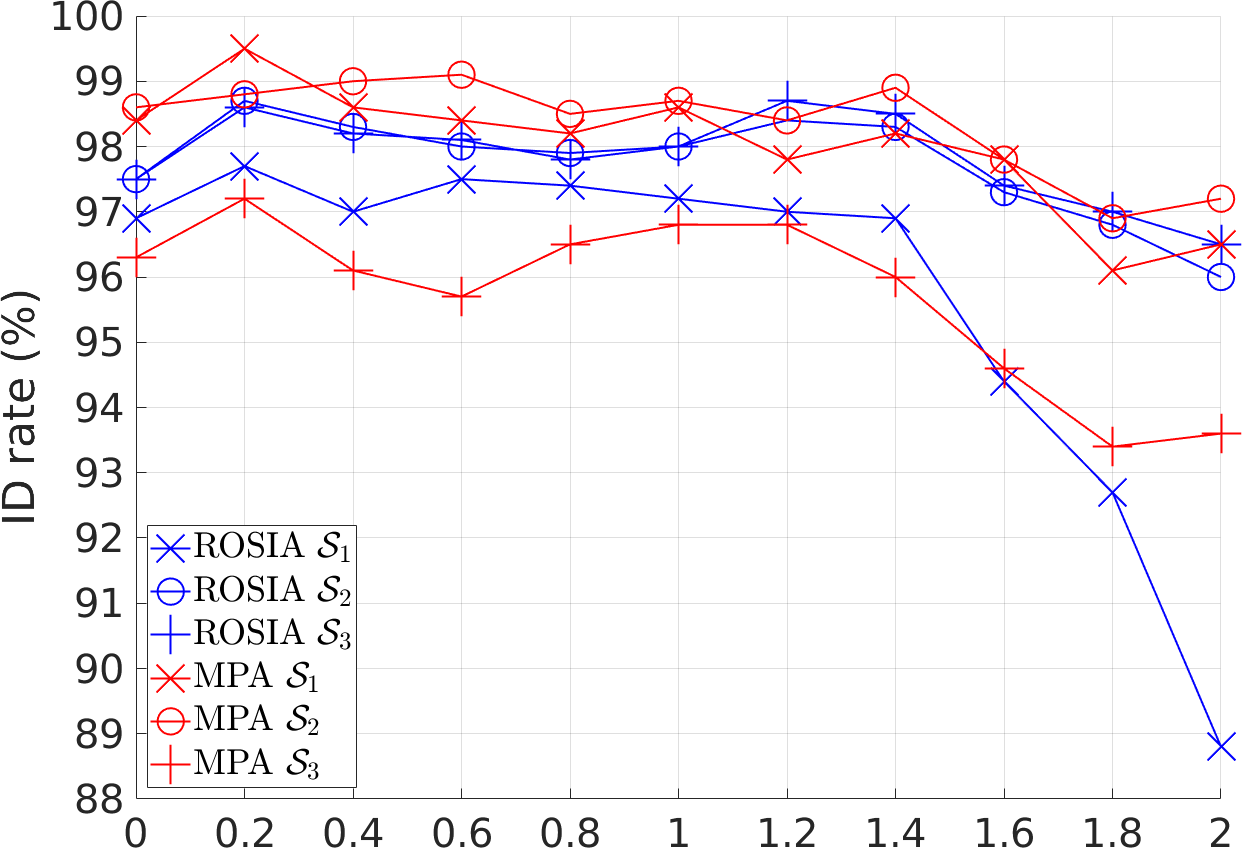}}\hspace{0.1em}
    \subfloat{\includegraphics[width=0.32\textwidth]{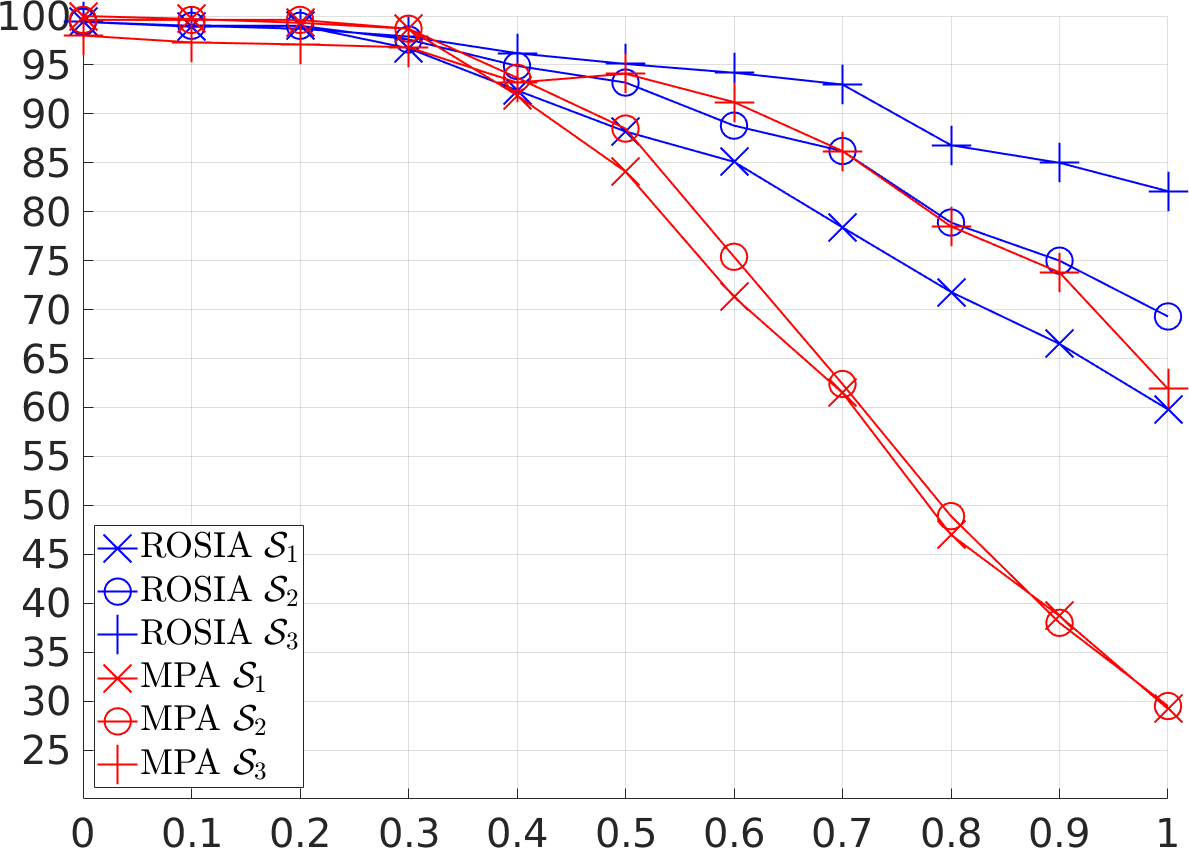}}\hspace{0.1em}
    \subfloat{\includegraphics[width=0.32\textwidth]{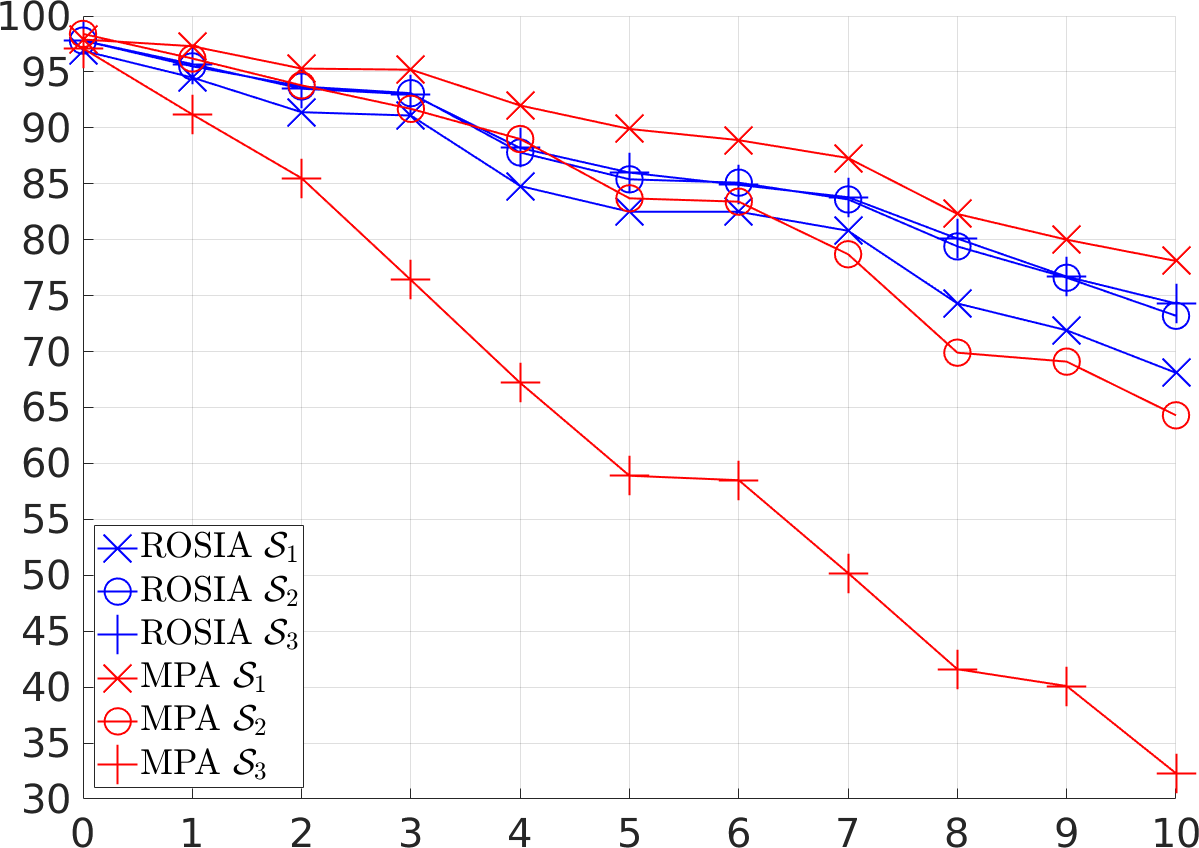}}\\
    \subfloat{\includegraphics[width=0.32\textwidth]{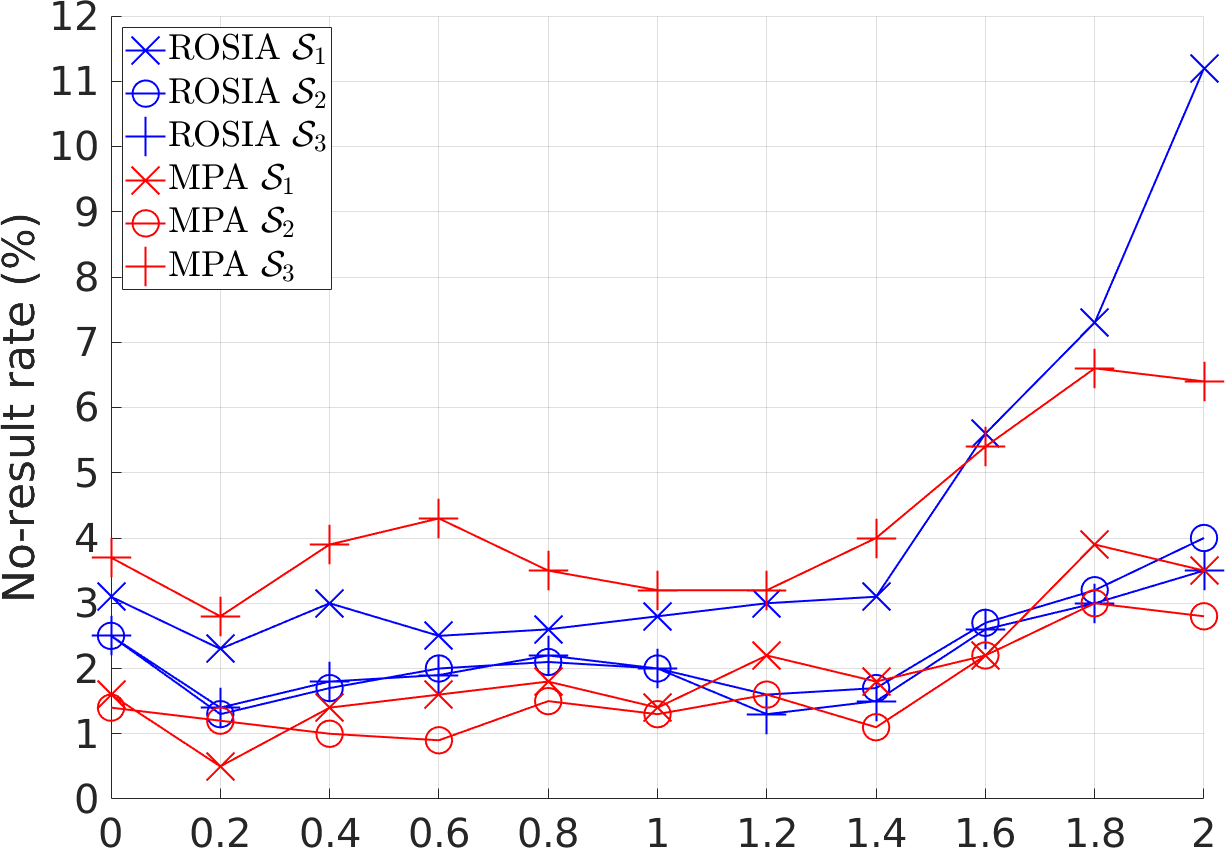}}\hspace{0.1em}
    \subfloat{\includegraphics[width=0.32\textwidth]{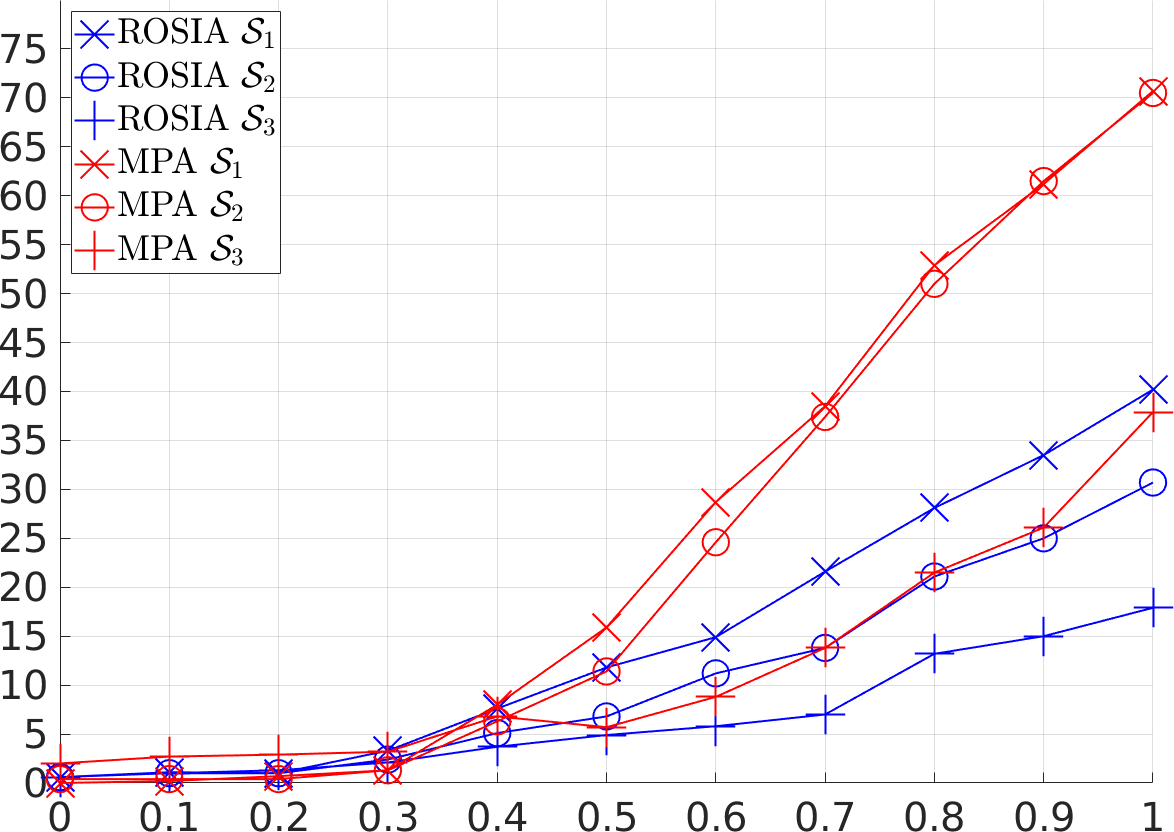}}\hspace{0.1em}
    \subfloat{\includegraphics[width=0.32\textwidth]{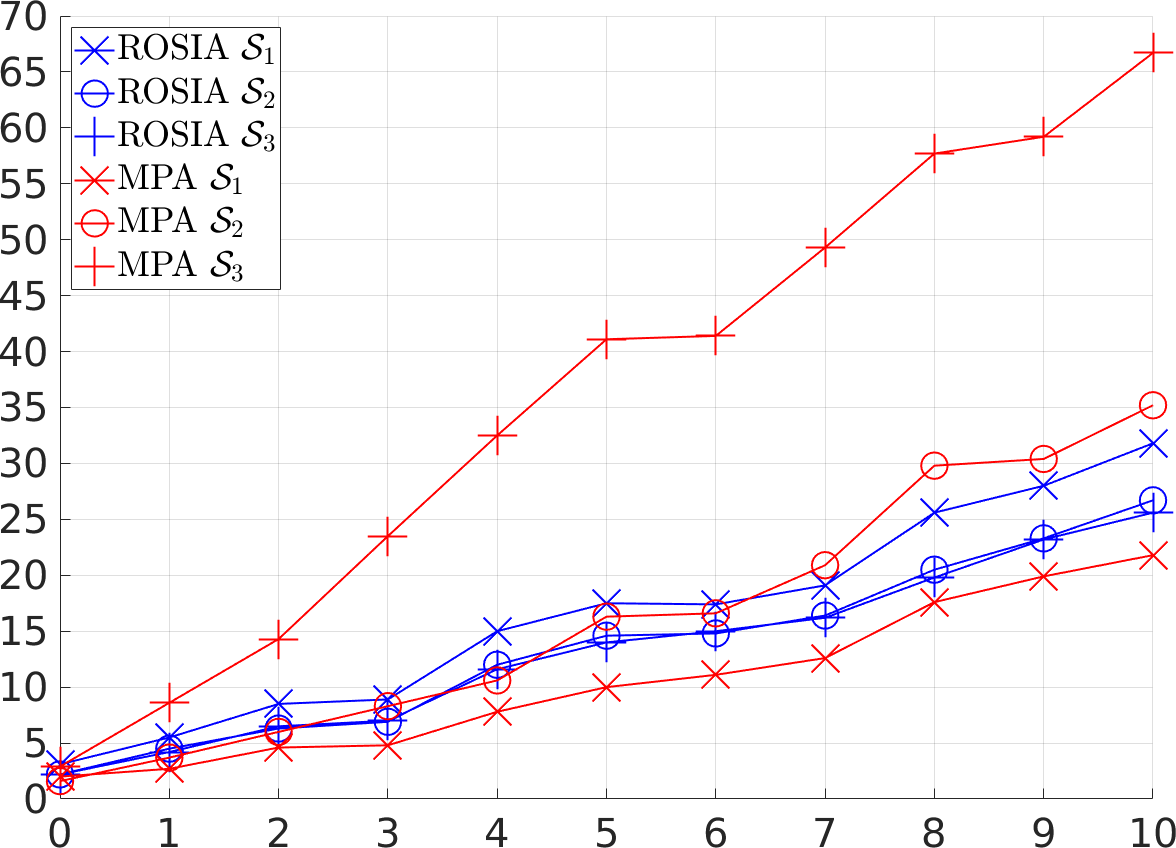}}\\
	\subfloat{\includegraphics[width=0.32\textwidth]{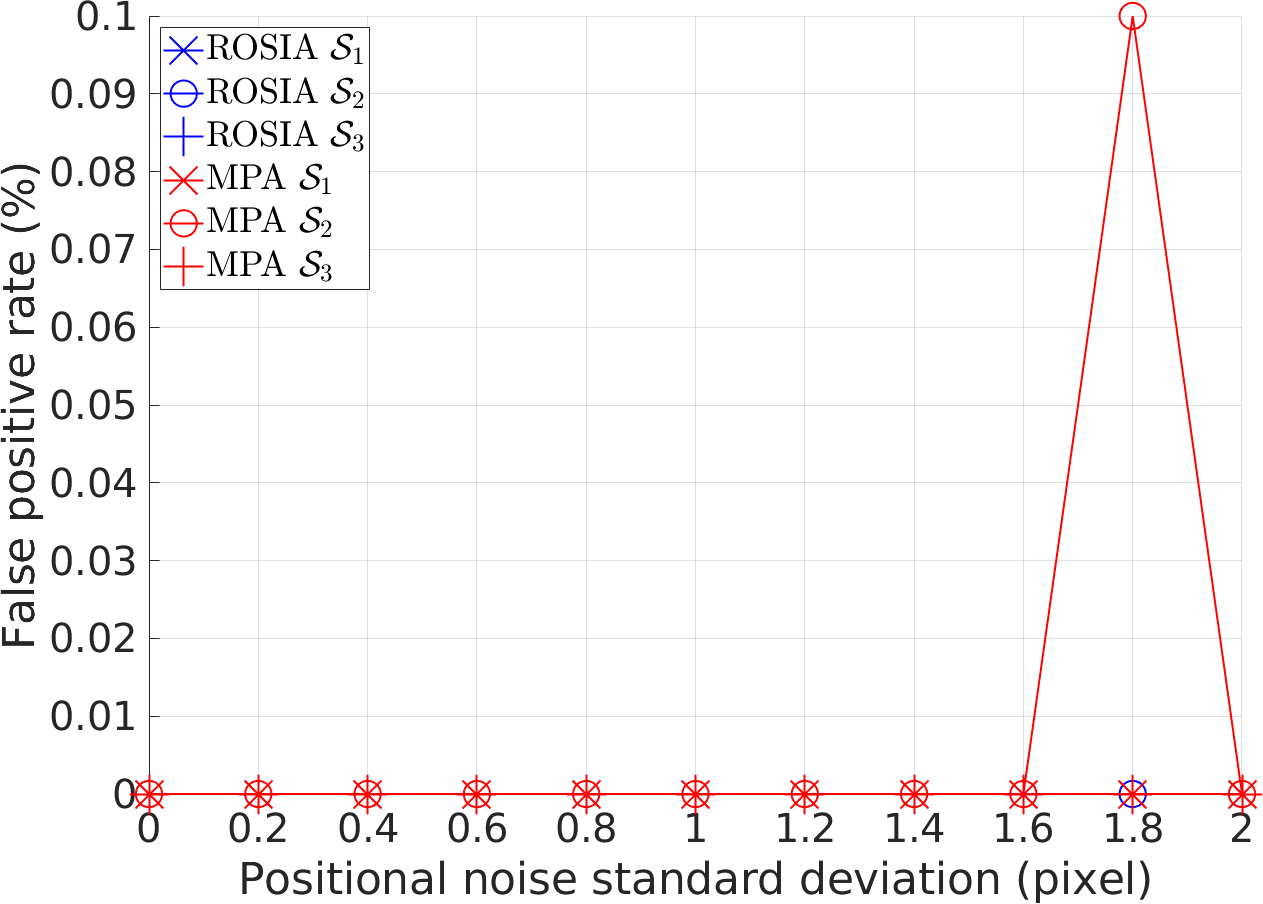}}\hspace{0.1em}
	\subfloat{\includegraphics[width=0.32\textwidth]{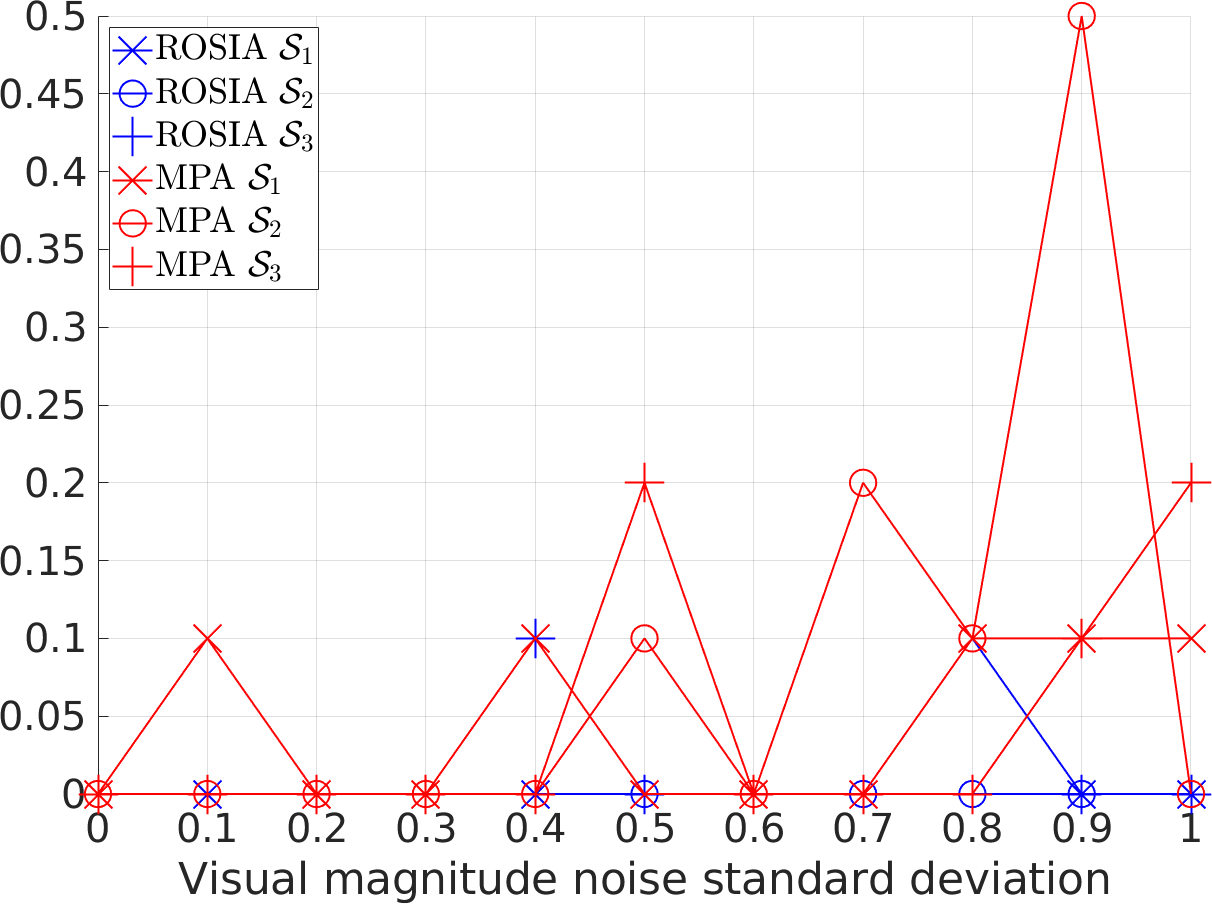}}\hspace{0.1em}
	\subfloat{\includegraphics[width=0.32\textwidth]{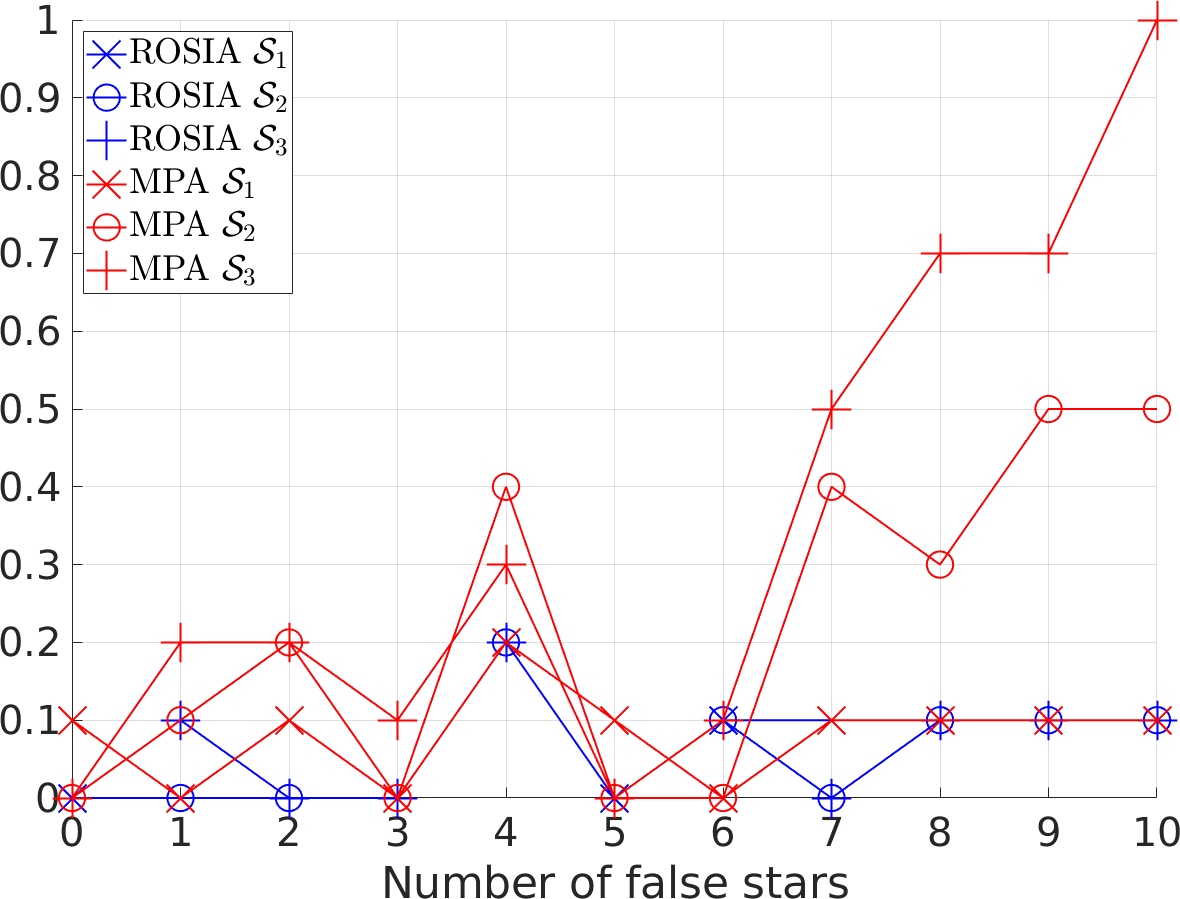}}\\
	\caption{Performance of ROSIA (in blue) and MPA (in red) against different sources of noise. Top row: ID rate. Middle row: No-result rate. Bottom row: false positive rate. $\mathcal{S}_1$, $\mathcal{S}_2$, and $\mathcal{S}_3$ denote three hyperparameters configurations. $\mathcal{S}_1$($-x-$) and $\mathcal{S}_2$($-\circ-$) are set at $1.5\sigma$ and $2\sigma$ of the standard pixel and visual magnitude noise, i.e., 1 pixel and 0.3 magnitudes. $\mathcal{S}_3$($-+-$) covers the maximum (pixel and magnitude) noise SDs.}\label{fig:exp_perf}\vspace{-0.2cm}
\end{figure*}

The performances of ROSIA and MPA with three configurations ($\mathcal{S}_1$, $\mathcal{S}_2$, and $\mathcal{S}_3$) can be seen in the first column of Fig. \ref{fig:exp_perf}. The top figure illustrates the ID rates against the positional noise. The general trends of ROSIA are expected - ROSIA $\mathcal{S}_3$ yields the best ID rates since its thresholds cover the noise SD sweeps, followed by ROSIA $\mathcal{S}_2$ and then ROSIA $\mathcal{S}_1$. The gap between the ID rates of ROSIA is small and stable, ranging from 96.9\% to 98.7\%, when the positional noise SD is smaller than 1.5 pixels. At 2 pixels (noise SD), ROSIA $\mathcal{S}_1$ drops to 88.8\%, while ROSIA $\mathcal{S}_3$ remains at 96.5\%. This is expected as the angular deviation threshold of $\mathcal{S}_1$ covers only potential star matches within 1.5 pixels of deviation.

MPA has a similar range of ID rates, varying from $93.4\%$ to $99.5\%$. Interestingly, MPA $\mathcal{S}_3$, the configuration that is least likely to drop out potential star matches, performs the worst. The main reason for this is that MPA often chooses the wrong pole stars due to ambiguity when many potential matches are considered. This phenomenon is much more significant in the false star experiment, where we elaborate on the root cause. 

The middle and bottom rows of the first column in Fig.~\ref{fig:exp_perf} plot the no-result and false positive rates, respectively. As we can see, none of the failure cases of ROSIA is false positive in this experiment. On the other hand, there is one false positive instance for MPA, which corresponds to $0.1\%$ (1/1000) when the noise SD is 1.8 pixels in the figure. 

\begin{figure*}[!t]
    \centering
    \subfloat{\includegraphics[width=0.32\textwidth]{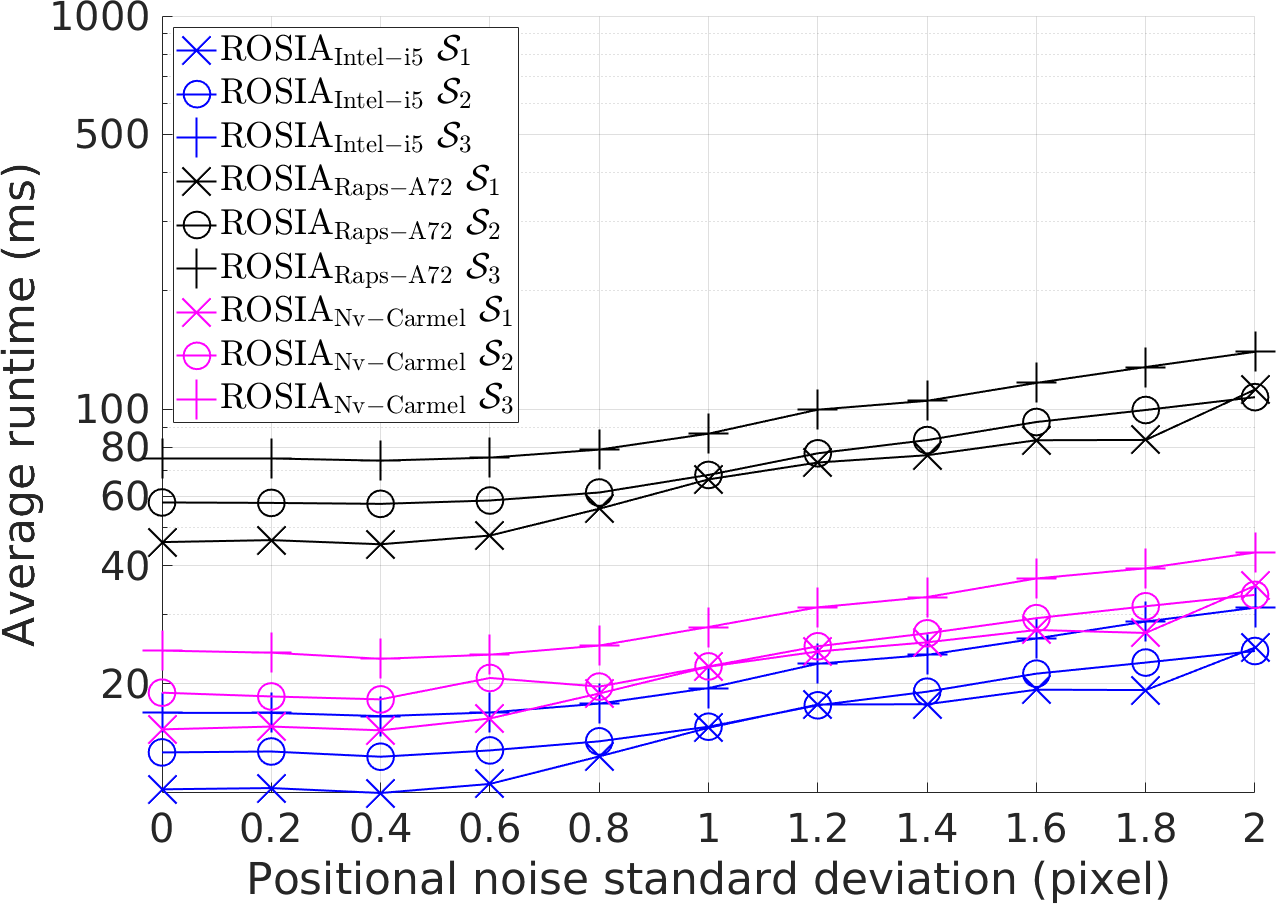}}\hspace{0.1em}
	\subfloat{\includegraphics[width=0.32\textwidth]{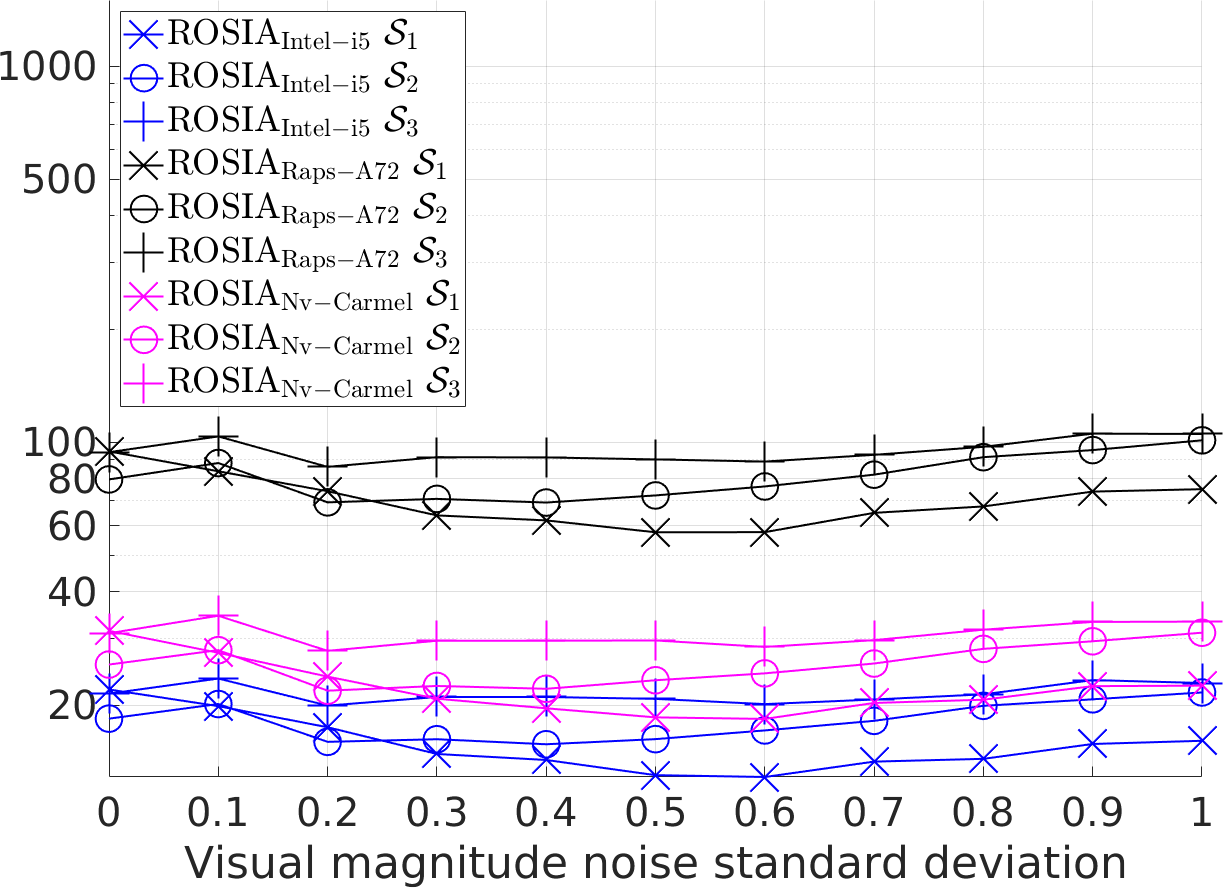}}\hspace{0.1em}
	\subfloat{\includegraphics[width=0.32\textwidth]{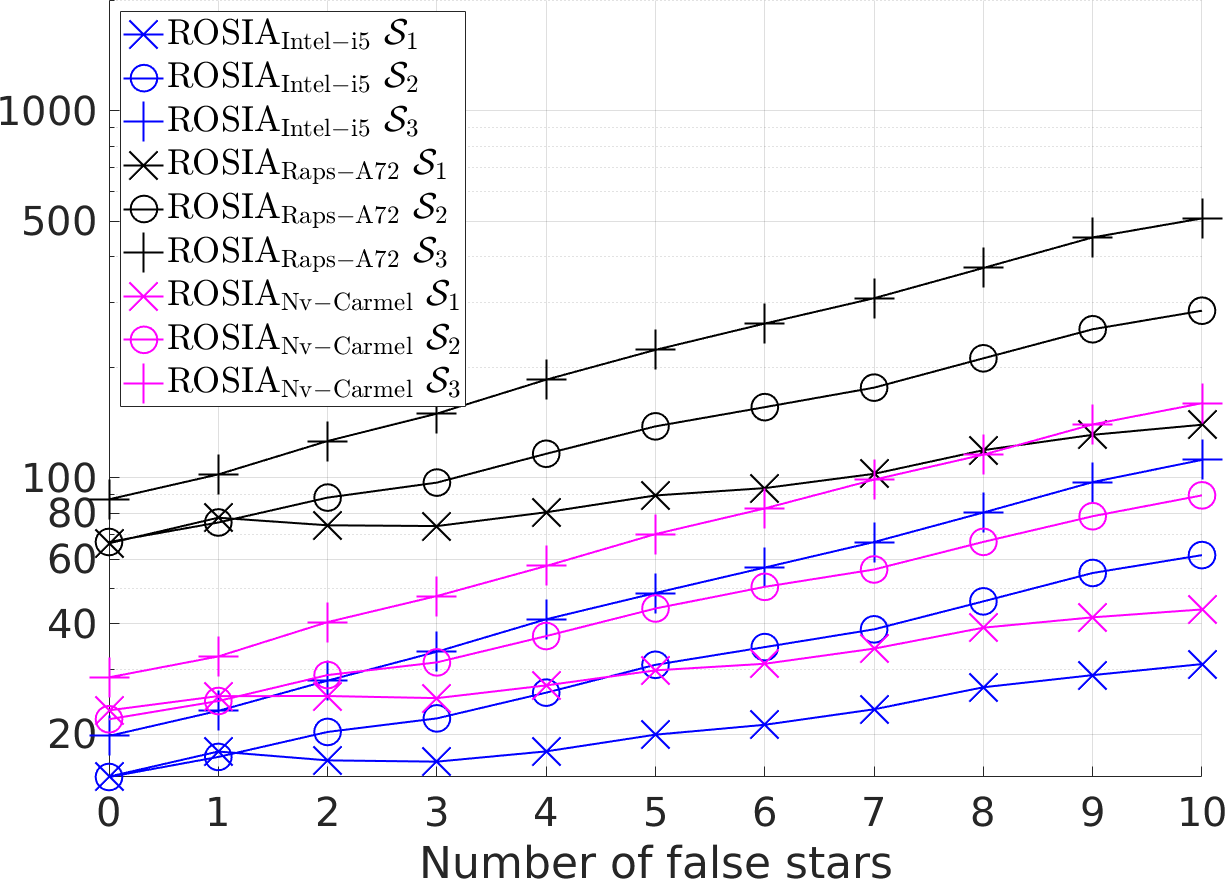}}\\
	\caption{The average runtime of ROSIA. Note the log scale of the y-axis. Consistent notations with Fig. \ref{fig:exp_perf}. The subscripts `Intel-i5', `Raps-A72', and `Nv-Carmel' differentiates the processor types used in our experiments.}\label{fig:exp_runtime}\vspace{-0.2cm}
\end{figure*}

Fig.~\ref{fig:exp_runtime} (left) plots the average runtimes of ROSIA\footnote{We did not evaluate the runtime of MPA since our implementation is not optimized in efficiency. } in this experiment. As expected, larger thresholds demand higher runtimes since the cardinality of each sub-catalog ($M_{sub}$) is larger. Besides, as the noise level increases, potential matches are more likely to be dropped, which harms the uniqueness of the query star pattern. Consequently, ROSIA has to visit a larger amount of the rotation space to identify the optimal solution. 

The average runtime of ROSIA increases as the noise level increases. We highlight the runtime of $\mathcal{S}2$ since it provides the optimal balance between ID rate and runtime. On i5, the average runtime rises from 13.39ms to 24.25ms. On A72, it grows from 57.9ms to 107.29ms, and on Carmel, it increases from 19ms to 33.7ms.\\

% Positional noise and deviation thresholds affect the definition of a match. For instance, a larger positional deviation threshold allows a looser matching definition; see the first term of our objective function~\eqref{eqn:Q_sid}. Consequently, the rotation's accuracy is impacted, which can be seen in Fig. \ref{fig:rot_err_exp}. At 1 pixel noise, the range of the error is $[\ang{0.066},\, \ang{0.081}]$; whereas at 2 pixel noise, it becomes $[\ang{0.096},\, \ang{0.105}]$. We emphasize that the accuracy of the rotation is not the priority of this work. 

\textbf{Visual magnitude noise.} In this experiment, we evaluate the robustness of both algorithms against visual magnitude noise. We fixed the positional SD at $0.0135^\circ$ ($1$ pixel) and injected zero-mean Gaussian noise with the SD ranging from $0$ to $1$ visual magnitude. The second column of Fig.~\ref{fig:exp_perf} plots the performance results.

The ID rates of the best-performing configuration of ROSIA (and MPA) at $0$, $0.5$, and $1.0$ noise levels are $99.4\%$ ($100\%$) to $95.1\%$ ($94.1\%$) and $82.1\%$ ($61.9\%$). In contrast to the positional noise experiment, both methods follow the same ranking pattern in terms of ID rate, i.e.,  $\mathcal{S}_3 > \mathcal{S}_2 > \mathcal{S}_1$. As previously alluded to, this is expected due to the coverage of the noise SD. The ID rates of MPA $\mathcal{S}_1$ and MPA $\mathcal{S}_2$ drop significantly as the noise increases - both at approximately $29.5\%$ when the noise SD is $1$. Meanwhile, the ID rates of ROSIA remain at $69.3\%$ ($\mathcal{S}_2$) and $59.8\%$ ($\mathcal{S}_1$).

We found that MPA suffers from finding a pole star candidate that satisfies the condition of having three (matched) neighbor stars within the allowed $Rp$ counts.
As the noise SD exceeds the deviation thresholds in $\mathcal{S}_1$ and $\mathcal{S}_2$, the dropout rate of potential matches (neighbors in this context) increases, which makes it hard to achieve the said condition. Numerically, when the visual magnitude noise SD is $1$, $26\%$ of the failure cases for MPA $\mathcal{S}_1$ stems from the failure of obtaining even the first pole star candidate. When MPA manages to find the first pole star candidate, it fails to find the second pole star candidate. The superiority of the top-down approach of ROSIA is illustrated here. It avoids the multiplicative nature of failure likelihoods in a bottom-up pipeline. 

Similar to both algorithms, most of the failure cases are no-result, as depicted in the middle and bottom rows of the second column in Fig.~\ref{fig:exp_perf}. The total false positive cases remain low for both: $2$ for ROSIA and $19$ for MPA with all three configurations in all $11000$ test instances.

There is no observable increasing pattern of the average runtime as the noise increases (middle plot in Fig.~\ref{fig:exp_runtime}). That is because the average number of detected stars decreases as the visual magnitude noise increases due to the exceeding of the visual magnitude detection threshold. Concretely, there are only $17$ stars on average when the visual magnitude noise SD is $1$, which on average has six fewer stars than when there is no visual magnitude noise. The smaller number of stars compensates for the challenge induced by the increment of noise, which leads to similar runtimes. The (average) average runtimes of the best ID rate configuration ($\mathcal{S}3$) are 21.47ms, 95.07ms, and 30.85ms on i5, A72, and Carmel, respectively.\\

\textbf{False stars.} Lastly, we evaluate the robustness of both algorithms against false stars. The number of false stars added to each testing image is swept from 0 to 10, corresponding to $0\%$ to $30\%$ of the total number of detected stars. Both the standard positional (1 pixel) and magnitude (0.3) SDs were applied in this experiment. 

The robustness of MPA against false stars is displayed in the third column of Fig.~\ref{fig:exp_perf}. For properly tuned thresholds, MPA $\mathcal{S}_1$ outperforms all combinations in terms of ID rates in all noise levels. The highest ID rates of ROSIA (and MPA) at $0$, $5$, and $10$ false stars are $97.8\%$ ($98.4\%$), $86\%$ ($89.9\%$), and $74.3\%$ ($78.1\%$), respectively. 

On the other hand, the robustness of ROSIA against ill-defined thresholds is portrayed by the small gaps between all three configurations. Specifically, ROSIA $\mathcal{S}_2$ and ROSIA $\mathcal{S}_3$ have approximately the same ID rates in all test cases since both (thresholds) cover at least twice the noise SDs. In contrast, the deviation of thresholds from the noise SDs tremendously affects MPA $\mathcal{S}_3$. In numbers, the lowest ID rates of ROSIA (and MPA) are $96.9\%$ ($97.1\%$), $82.5\%$ ($58.9\%$), and $68.1\%$ ($32.3\%$), which correspond to $0$, $5$, and $10$ false stars, respectively. 

We associate this with the difference between MPA and ROSIA in terms of star representation. The pole star representation of MPA ignores the angular distances between the neighboring stars. The lack of such constraint allows a wrong pole star to be selected, especially with MPA $\mathcal{S}_3$ due to its looser thresholds. On the contrary, ROSIA compensates for the poorly tuned thresholds with an optimal representation - the complete graph which contains maximum input information.

The vast majority of the failure cases are no-result, as depicted in the middle and bottom rows of the third column in Fig.~\ref{fig:exp_perf}. In total, there are only $16$ false positive cases for ROSIA and $62$ cases for MPA.

Similar to the positional noise experiments, the average runtime of ROSIA increases as the number of false stars increases, as seen in the right plot of Fig.~\ref{fig:exp_runtime}. Again, we highlight the runtime of $\mathcal{S}_2$ as it provides the best trade-off between ID rate and runtime. On i5, the average runtime rises from 15.3ms to 61.55ms. On A72, it grows from 66.83ms to 284.9ms, and on Carmel, it increases from 21.9ms to 89.53ms.

This experiment also highlights the importance of selecting the optimal thresholds. In the case of ROSIA $\mathcal{S}_3$, setting a threshold value higher than the measurement uncertainty (magnitude noise SD = 0.3 and $\epsilon_v = 1.2$) does not impact the ID rate but increases the number of evaluations, which in turn harms the runtime performance.\\

\textbf{Memory consumption.} We analyze the memory consumption of both methods in this section. The average memory consumption of ROSIA and MPA in all experiments is 3MB and 7MB, respectively. Additionally, ROSIA is demonstrated to be more efficient in terms of up-scaling. The size of the onboard catalog size depends on the camera FOV and magnitude threshold, which vary depending on the application setting. As discussed in Sec.~\ref{sec:lit_review}, the storage complexity of MPA grows quadratically with the size of the input data ($M$), i.e., $\mathcal{O}(M^2)$ with a limitation incorporated by the camera FOV. On the contrary, ROSIA has a linear growth rate of $\mathcal{O}(M)$.

\begin{figure*}[!t]
    \centering
    \subfloat{\includegraphics[width=0.42\textwidth]{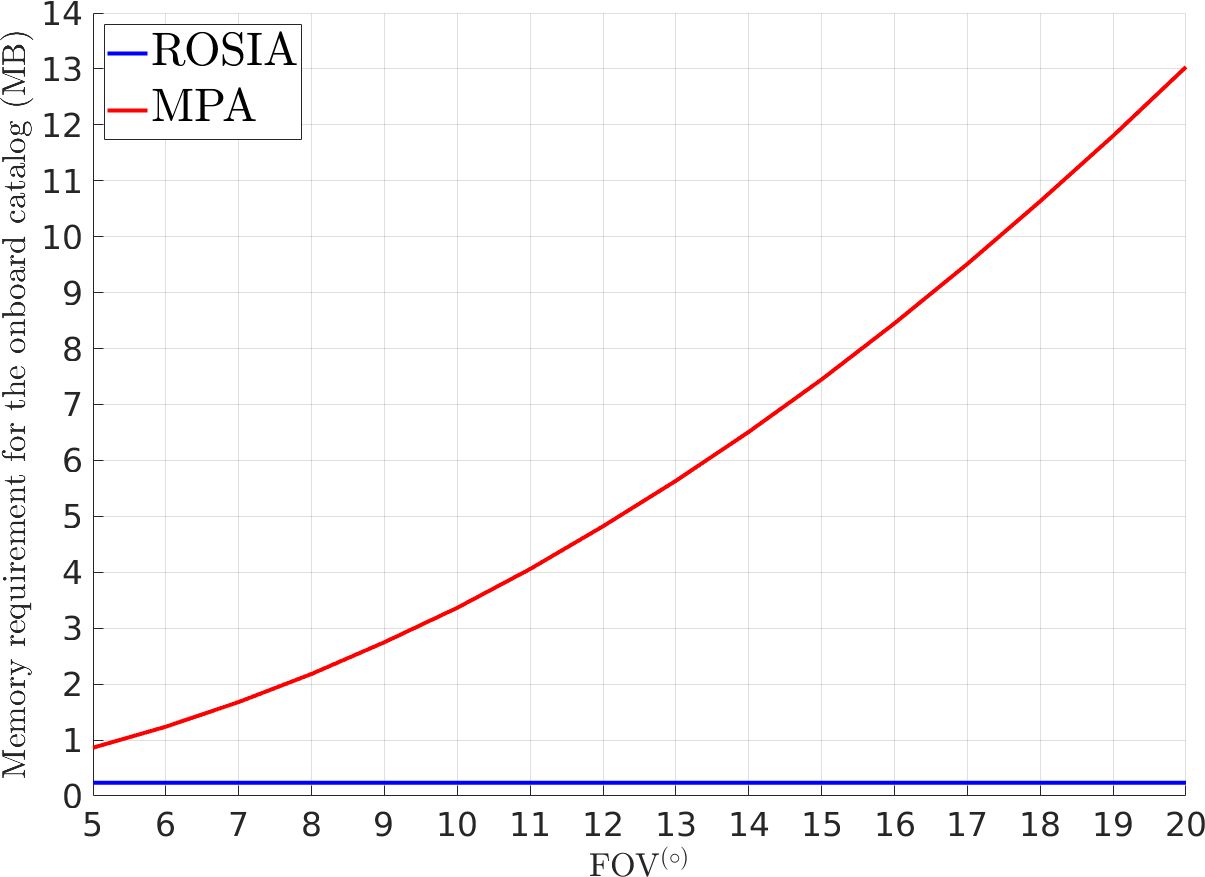}}
    \subfloat{\includegraphics[width=0.42\textwidth]{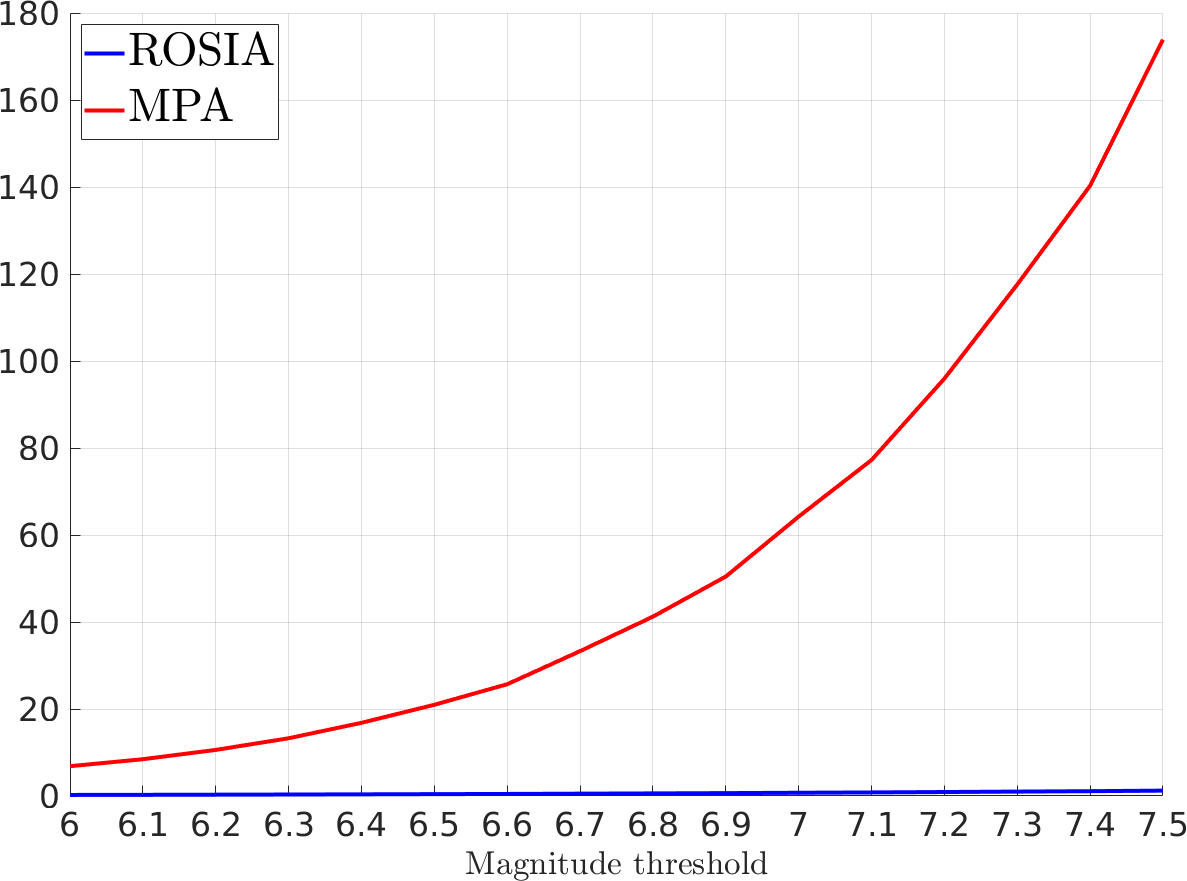}}
	\caption{Memory requirement of ROSIA and MPA for the onboard catalog against different camera FOVs and visual magnitude thresholds.}\vspace{-0.2cm}
\label{fig:mem_con}
\end{figure*}

We plot the memory footprint of both methods against the camera FOV and visual magnitude threshold in Fig.~\ref{fig:mem_con}. The left plot of Fig.~\ref{fig:mem_con} plots the memory footprint in relation to the camera FOV. The visual magnitude threshold was fixed at 6, which was used in the experiments above. As the FOV increases, the memory consumption of MPA increases as well since it relies on the FOV to limit the number of star-pair combinations during the onboard catalog construction process. The memory consumption ranges from 0.8MB at a 5 FOV to 13MB at a 20 FOV. On the other hand, the memory footprint of ROSIA remains constant at approximately 0.24MB, as the FOV does not affect the total number of onboard catalog stars.

The right plot of Fig.~\ref{fig:mem_con} shows the memory footprints in relation to the visual magnitude threshold. The visual magnitude threshold represents the limit for the brightness of the stars that are included in the onboard catalog. When the magnitude threshold is increased from 6 to 7.5, the number of stars increases from 4934 to 25713, resulting in an increase in memory consumption for both methods. However, the increase in the memory footprint of ROSIA is linear, ranging from 0.24MB to 1.23MB, while the increase in the memory footprint of MPA is much more significant, ranging from 6.8MB to 173.9MB.

\subsection{Real data experiments}
In this experiment, we used ten real star images obtained from two public sources: the NASA library~\cite{nasalibrary} and Astrometry~\cite{Astrometry}. We show four examples in this section and the rest in Sec.~\ref{Sec:AppendixC}. The constellations captured in these images are summarized in Tab.~\ref{tab:real_data}. We use Astrometry.net \cite{lang2010astrometry} to perform calibration, star detection, and centroiding. Then, we back-projected the top 50 brightest pixels to body-vectors using the estimated camera parameters. Since our onboard catalog has only stars with magnitude six or lower, we filter the input stars accordingly, and the number of input body-vectors to ROSIA is summarized in Tab.~\ref{tab:real_data}.

MPA and ROSIA share similar identification results in these images, as seen in Fig.\ref{fig:real_data_1} and \ref{fig:real_data_2} (more in Fig.~1, 2, and 3 in Sec.~\ref{Sec:AppendixC}). The identified stars are annotated with their index in Fig.\ref{fig:real_data_1} and \ref{fig:real_data_2}. The average angular deviation between the body-vectors and their corresponding inertial-vectors can be seen in Tab.~\ref{tab:real_data}. These errors are caused by the sub-optimal camera parameters estimation from Astrometry.net. We highlight that these errors are higher than the usual measurement uncertainty in well-calibrated star tracker systems (in the range of arcseconds). ROSIA's runtimes are summarized in Tab.~\ref{tab:real_data}, ranging from 6.6ms to 57.3ms on i5, 50.6ms to 826ms on A72, and 18.3ms to 279ms on Carmel.

% To combat this, we set $\alpha_\epsilon = 0.052^{\circ}$ in all runs. 

% In this experiment, we used four real star images obtained from the NASA library\cite{nasalibrary}. The constellations captured in these images are summarized in Tab.~\ref{tab:real_data}. We first use Astrometry.net \cite{lang2010astrometry} to perform calibration, star detection and centroiding. Then, we back-projected the top 50 brightest pixels to body-vectors using the estimated camera parameters. Since our onboard catalog has only stars with magnitude six or lower, we filter the input stars accordingly, and the number of input body-vectors to ROSIA are summarized in Tab.~\ref{tab:real_data}.

% ROSIA was able to successfully identify at least seven stars in these images, indicating that the Star-ID task was successful. The identified stars are annotated with their index in Fig.\ref{fig:real_data_1} and \ref{fig:real_data_2}. The average angular deviation between the body-vectors and their corresponding inertial-vectors can be seen in Tab.~\ref{tab:real_data}. These errors are caused by the sub-optimal camera parameters estimation from Astrometry.net. We highlight that these errors are higher than the usual measurement uncertainty in well-calibrated star tracker systems (in the range of arcseconds). To combat this, we set $\alpha_\epsilon = 0.052^{\circ}$ in all runs. The runtimes of these runs are summarized in Tab.~\ref{tab:real_data}, ranging from 6.6ms to 57.3ms on i5, 50.6ms to 826ms on Raps-A75, and 18.3ms to 279ms on Nv-Carmel.

\begin{figure}[!t]
    \centering
    \subfloat{\includegraphics[width=0.45\textwidth]{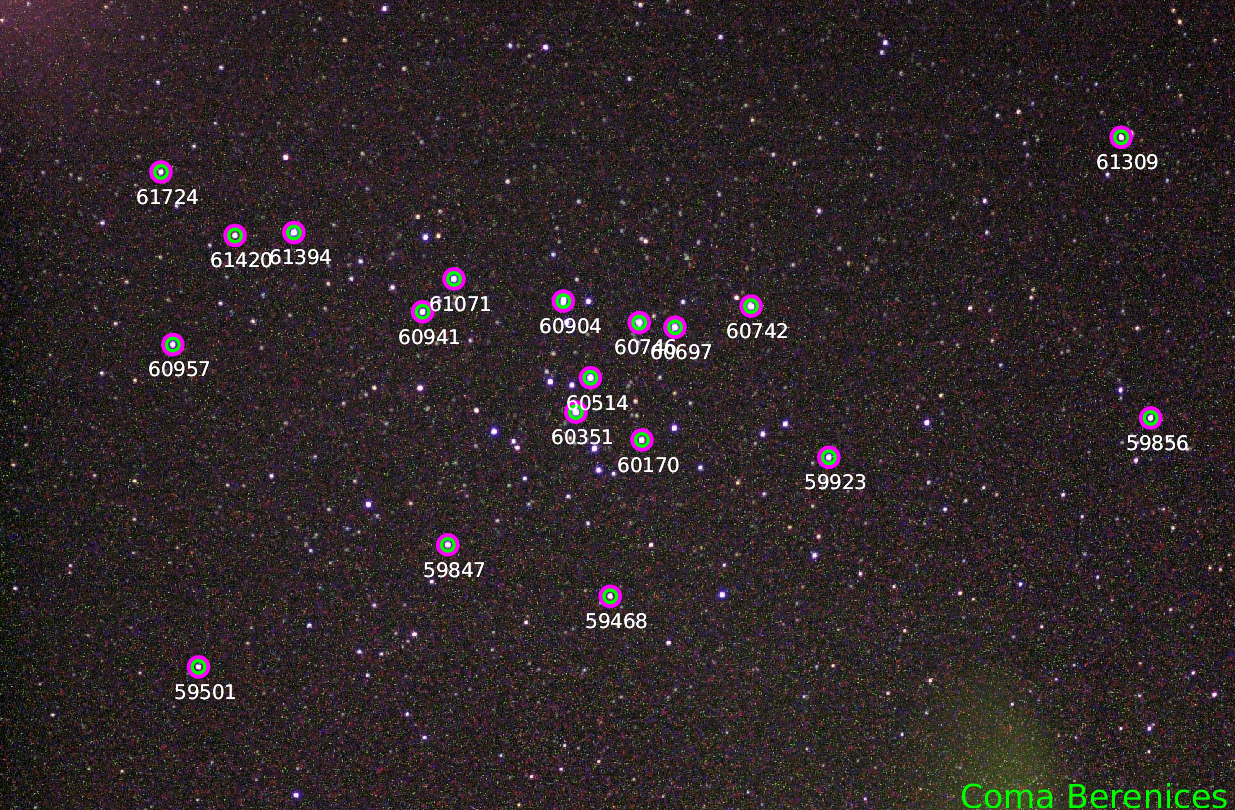}}\\
    \subfloat{\includegraphics[width=0.45\textwidth]{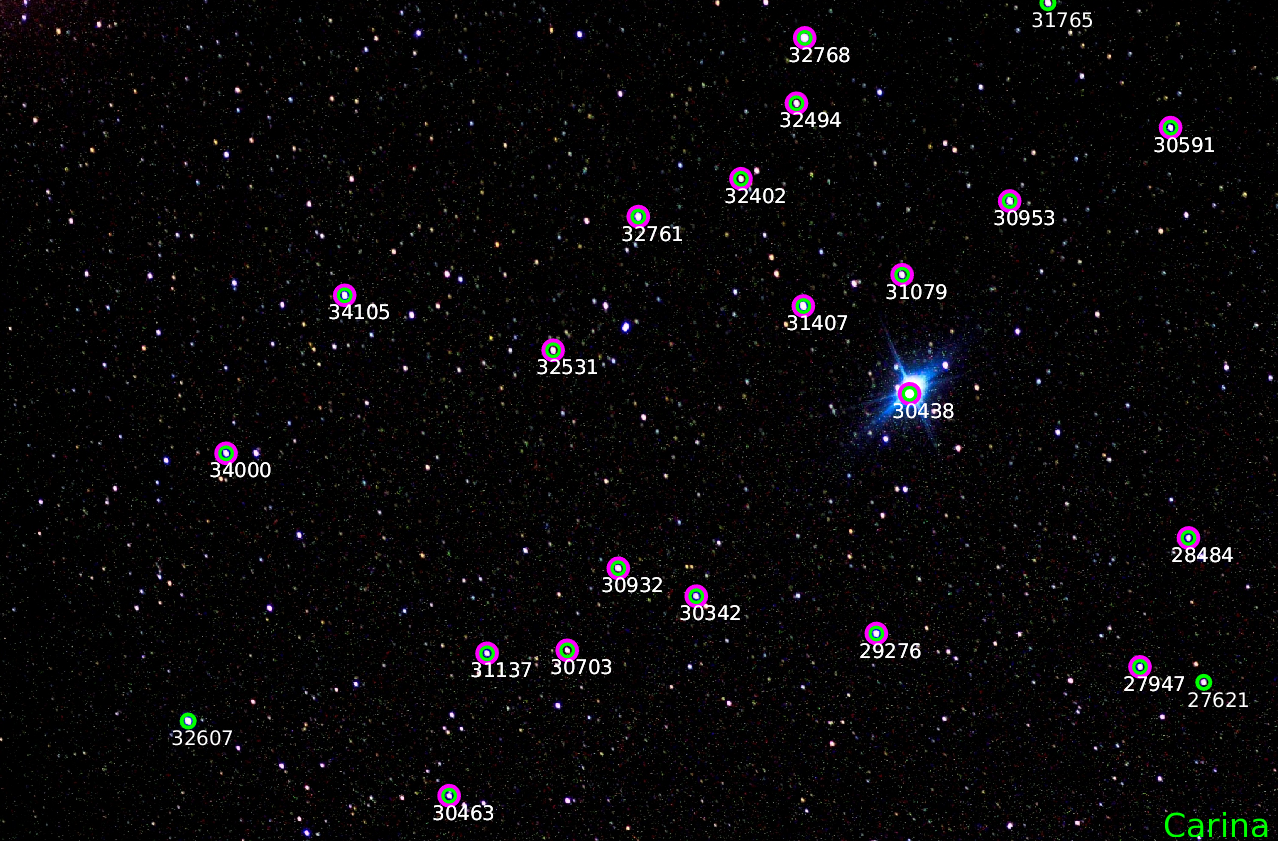}}\\
	\caption{Identification results of ROSIA (green circles) and MPA (magenta circles) on real star images. The constellation in each image is annotated at the right bottom part of the image. The identified stars are annotated with their indexes in the Hipparcos catalog.}\vspace{-0.5cm}
\label{fig:real_data_1}
\end{figure}

\begin{figure}[!t]
    \centering
    \subfloat{\includegraphics[width=0.45\textwidth]{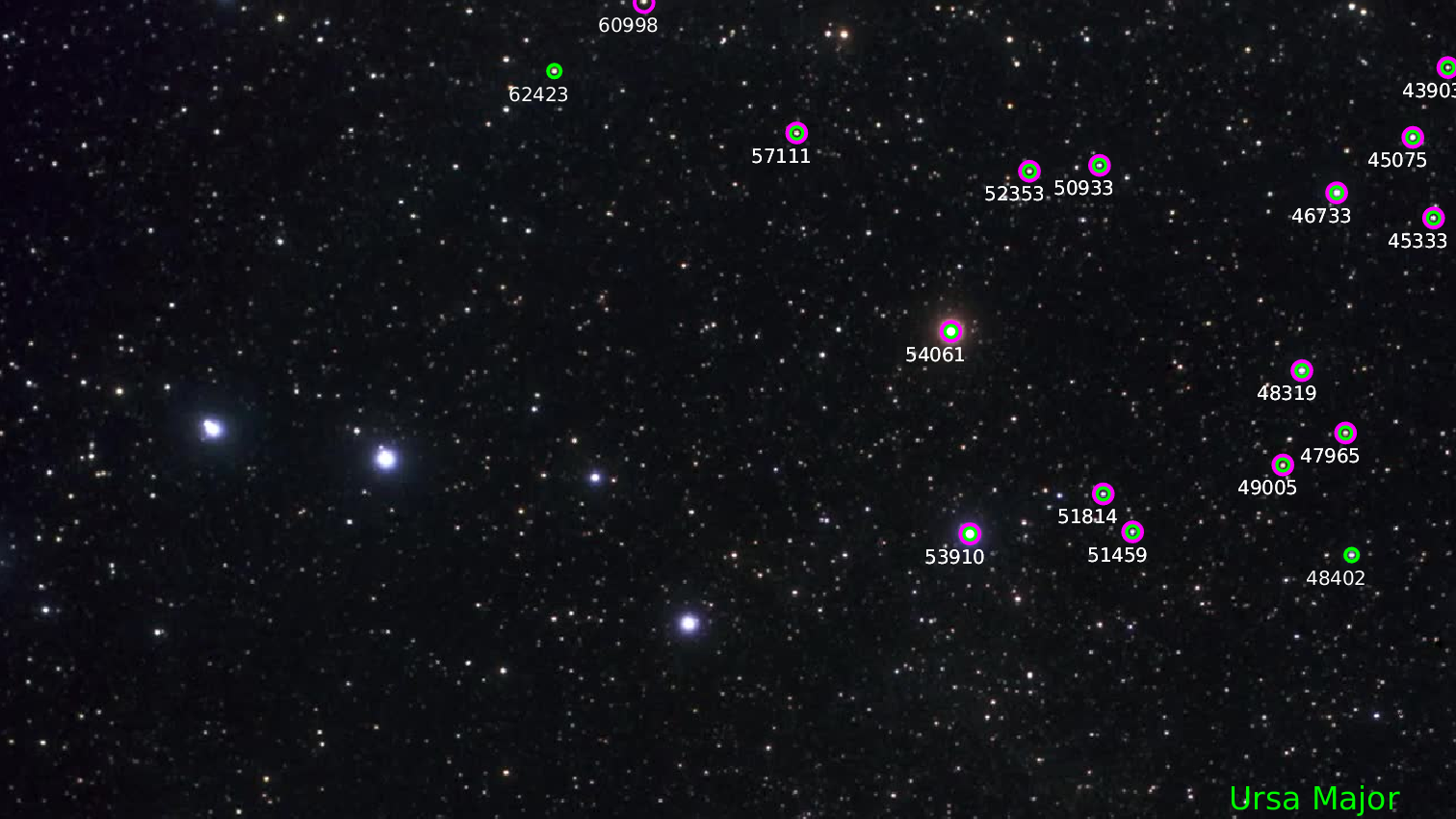}}\\
    \subfloat{\includegraphics[width=0.45\textwidth]{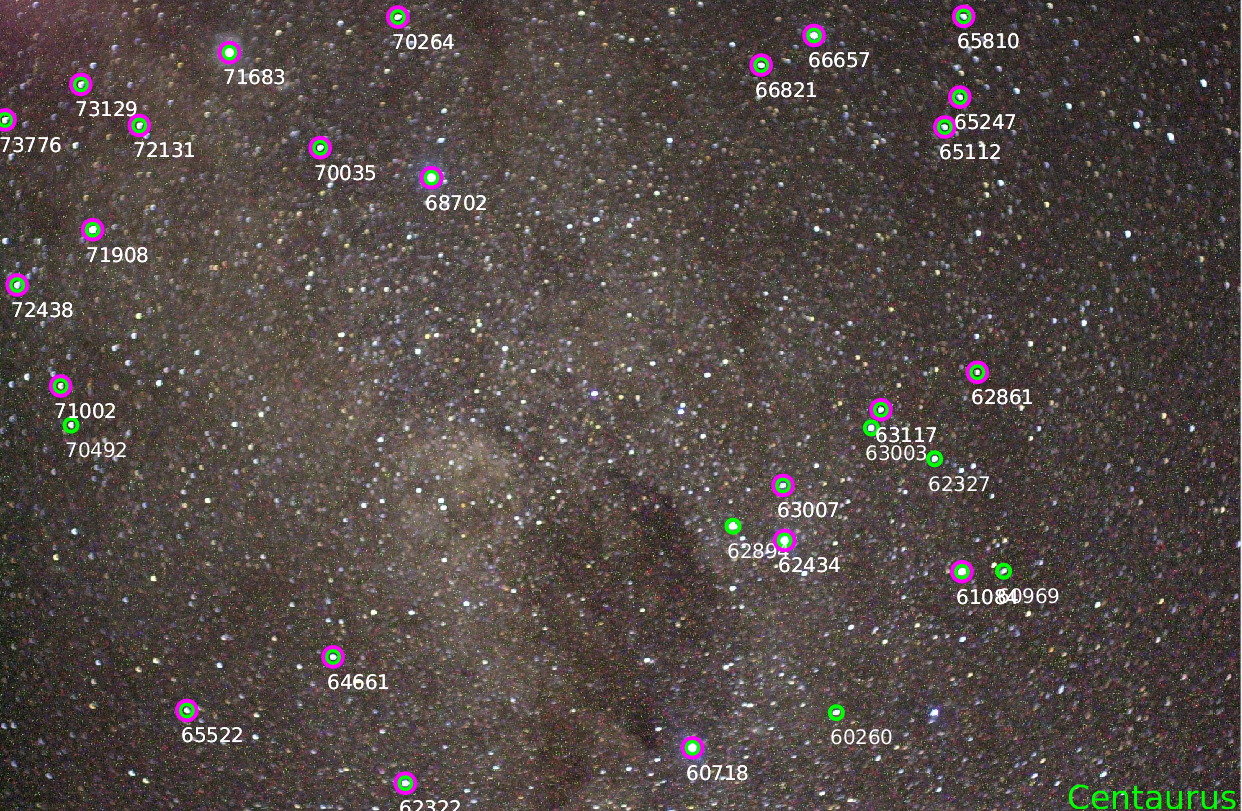}}\\
	\caption{Continuation of Fig.~\ref{fig:real_data_1}}\vspace{-0.5cm}
\label{fig:real_data_2}
\end{figure}

\begin{table}
	\begin{center}
        \centering
		\caption{Details of the real star images and runtime results. The angle deviation is the average of all inputs.}\label{tab:real_data}
		\begin{tabular}{cccccc}
			\toprule
          \multirow{2}{*}{Constellation} & \multirow{2}{*}{input} & \multirow{2}{*}{Angle} & \multicolumn{3}{c}{Runtime (ms)}\\
          \cmidrule(lr){4-6}
           & {count} &  {deviation($\circ$)}& i5 & A72 & Carmel \\
          \cmidrule(lr){1-6}
            Coma Berenices & 19 & 0.01 & 6.6 & 50.6 & 18.3  \\
            Carina & 23 & 0.008 & 11.6 & 101.8 & 37.1 \\
            Ursa Major & 18 & 0.06 & 15.5 & 124.9 &44.2\\
			  Centaurus & 30 & 0.04 & 57.3 &826 &279\\
			\bottomrule
		\end{tabular}\vspace{-0.5cm}
	\end{center}
\end{table}

\section{Limitations}
We discuss the limitation of ROSIA and present a future direction here. We highlight that $Q_{\rm{ROSIA}}$ without the triplet constraint, i.e., $K = 0$, is the best objective function for the Star-ID task in terms of ID rates. In the presence of noise, the triplet constraint could potentially eliminate valid star matches. For instance, a match between the rotated scene and catalog star in terms of angular distance is only counted as a proper match if their two closest stars match. Such a likelihood is reduced in the presence of noise, especially the appearance of false stars.

However, the algorithm is not runtime feasible if the triplet constraint is removed, as we demonstrated in Sec.~\ref{sec:prob_formulation}. As such, we leave the search for a better formulation or potentially a better evaluation scheme in terms of the ID rate and runtime trade-off as future work.

\section{Conclusion}
We presented a new paradigm to solve the Star-ID problem. This paper shows that the seemingly computationally expensive rotation-search-based method, which was first conceptually formulated by Junkins et al. \cite{junkins1977star} but unresolved since then, can be implemented feasibly in a BnB fashion. The two main contributing factors are a tight upper-bound function and an efficient evaluation scheme. ROSIA achieves state-of-the-art performance in the presence of commonly evaluated noise sources. Besides, we show that it is robust against ill-tuned hyperparameters. In terms of feasibility, ROSIA runs in the range of milliseconds on CubeSat-qualified onboard computers. Lastly, we show that ROSIA is memory efficient and scales linearly with the size of the star catalog.

\bibliography{example.bib} % 
\bibliographystyle{IEEEtran}

\appendix
% \documentclass[12pt, draftclsnofoot, onecolumn]{IEEEtran}
% % \documentclass[a4paper]{article}
% \usepackage[english]{babel}
% \usepackage[utf8x]{inputenc}
% \usepackage{amsmath,amsfonts,amssymb}
% \usepackage{amsthm}
% \usepackage{graphicx, epstopdf}
% \usepackage[colorinlistoftodos]{todonotes}
% \usepackage{subfig}
% \usepackage{amsfonts}
% % \usepackage{footmisc}

% \usepackage{siunitx}
% \usepackage{algorithm}
% \usepackage{algorithmic}
% \usepackage[T1]{fontenc}
% \usepackage{mathtools}

% \usepackage{multirow}
% \usepackage{tabularx} % allow set table width
% \usepackage{booktabs}
% \usepackage{threeparttable}
% \usepackage{tablefootnote}

% \usepackage{xcolor}
% \usepackage{soul}

% new command section
% \newcommand{\R}{\textbf{R}}
% \newcommand{\Rmath}{$\textbf{R}$}

% \newcommand{\C}{\textbf{C}}

% defining vectors here
% \newcommand{\bs}{\textbf{s}}
% \newcommand{\bc}{\textbf{c}}
% \newcommand{\bu}{\textbf{u}}
% \newcommand{\bv}{\textbf{v}}
% \newcommand{\bx}{\textbf{x}}
% \newcommand{\bp}{\textbf{p}}

%\newcommand{\Ru}{$\textbf{R}_{\textit{\textbf{u}}}$}

% \newcommand{\Ru}{\R_{\bu}}
% \newcommand{\setC}{$\textbf{C}$\:}
% \newcommand{\setS}{$\textbf{S}$\:}
% \newcommand{\setM}{$\textbf{M}$\:}

% defining sub domain
% \newcommand{\B}{\mathbb{B}}

% % defining functions here
% \newcommand{\Qang}{Q_{\rm{ang}}}
% \newcommand{\Qros}{Q_{\rm{ROSIA}}}

% \newcommand{\uQang}{\overline{Q}_{\rm{ang}}}
% \newcommand{\uQros}{\overline{Q}_{\rm{ROSIA}}}

% \newtheorem{lem}{Lemma}{\bfseries}{\itshape}

\section{Supplementary materials for ROSIA: Rotation-Search-Based Star Identification Algorithm}

% \begin{document}
% \maketitle
\subsection{Necessary conditions of the upper bound and lower bound functions}\label{Sec:AppendixA}
In a maximization problem, the upper bound function of a BnB algorithm needs to fulfil the following condition,

\begin{equation}\label{eqn:cond1}
        \overline{Q}_{\rm{ROSIA}}(\mathbb{B}) \geq \max_{\mathbf{r} \in \mathbb{B}} {Q}_{\rm{ROSIA}}(\mathbf{R}_\mathbf{r}) \: .
\end{equation}

\noindent In words, the upper bound ($\overline{Q}_{\rm{ROSIA}}$) of any given domain $\mathbb{B}$ has to be larger or equal to the maximum objective value (${Q}_{\rm{ROSIA}}$) in $\mathbb{B}$. The lower bound ($\underline{Q}_{\rm{ROSIA}}$), on the other hand, has to fulfil the following condition,

\begin{equation}\label{eqn:cond2}
        \underline{Q}_{\rm{ROSIA}} \leq \max_{\mathbf{r} \in \omega} {Q}_{\rm{ROSIA}}(\mathbf{R}_\mathbf{r}) \: .
\end{equation}

\noindent In words, the lower bound only has to be lower or equal to the optimal objective value in the entire search domain $\omega$, which could be any sub-optimal ${Q}_{\rm{ROSIA}}$. 

In addition, both bounds are required to have the monotonicity property. The upper bound value $\overline{Q}_{\rm{ROSIA}}$ decreases monotonically as the domain gets branched to smaller sub-domains iteratively (see (7) in the manuscript). Meanwhile, the lower bound $\underline{Q}_{\rm{ROSIA}}$ is progressively updated to the current best ${Q}_{\rm{ROSIA}}$ (denoted as ${Q}*$ in Algorithm 1, line 10) in ROSIA.

\subsection{Proof that $\overline{Q}_{\rm{ROSIA}}$ is an upper bound to ${Q}_{\rm{ROSIA}}$ given any cube $\mathbb{B}$.}\label{Sec:AppendixB}
$\overline{Q}_{\rm{ROSIA}}$ has to fulfil the following conditions to qualify as a valid bounding function for BnB.
\begin{lem}
For any cube $\mathbb{B}$,
    \begin{equation}\label{eqn:cond1}
        \overline{Q}_{\rm{ROSIA}}(\mathbb{B}) \geq \max_{\mathbf{r} \in \mathbb{B}} {Q}_{\rm{ROSIA}}(\mathbf{R}_\mathbf{r}) \: .
    \end{equation}
\noindent When $\mathbb{B}$ collapses to a single point $\mathbf{r}$,
    \begin{equation}\label{eqn:cond2}
        \overline{Q}_{\rm{ROSIA}}(\mathbb{B}) = {Q}_{\rm{ROSIA}}(\mathbf{R}_\mathbf{r}) \: .
    \end{equation}
\begin{proof}
    To prove \eqref{eqn:cond1}, it is sufficient to prove that if a pair of (detected and catalog) stars, $\textbf{s}_i$ and $\textbf{c}_j$, contributes 1 to ${Q}_{\rm{ROSIA}}(\mathbf{R}_\textbf{r})$, i.e., $\angle (\mathbf{R}\textbf{s}_i, \textbf{c}_j) \leq \alpha_\epsilon \prod^{K}_{k} \lfloor | \theta_k^{(i)} - \phi_k^{(j)} | \leq 2\alpha_\epsilon \rfloor$, it must also contribute 1 to $\overline{Q}_{\rm{ROSIA}}(\mathbb{B})$. We first highlight that the triplet constraint, i.e., $\prod^{K}_{k} \lfloor | \theta_k^{(i)} - \phi_k^{(j)} | \leq 2\alpha_\epsilon \rfloor$, is not a function of rotation, hence can be omitted in the following proof. 
    
    Since $\textbf{r} \in \mathbb{B}$, hence, $\angle (\mathbf{R}_\textbf{u}\textbf{s}_i, \mathbf{R}_\textbf{r}\textbf{s}_i) \leq \alpha_{\mathbb{B}}$ based on 1) inequality (4) in the main paper and 2) the construction of $\mathbb{B} = \{\textbf{u}, \alpha_{\mathbb{B}}\}$, where $\textbf{u}$ is the center point of the cube $\mathbb{B}$, and $\alpha_{\mathbb{B}}$ is the distance of $\textbf{u}$ with respect to the furthest point in the cube. As such, $\angle (\mathbf{R}_\textbf{u}\textbf{s}_i, \textbf{c}_j) \leq \alpha_\epsilon + \alpha_{\mathbb{B}}$, hence, contributing 1 to $\overline{Q}_{\rm{ROSIA}}(\mathbb{B})$. 
    
    To prove \eqref{eqn:cond2}, see that when $\mathbb{B}$ collapses, $\textbf{r} = \textbf{u} = \textbf{v}$ and $\alpha_{\mathbb{B}} = 0$ (see (5) in the main paper), yielding \eqref{eqn:cond2}. 
\end{proof}
    
\end{lem}

\subsection{Real star images results}\label{Sec:AppendixC}

\begin{figure*}
    \centering
    \subfloat{\includegraphics[width=0.8\textwidth]{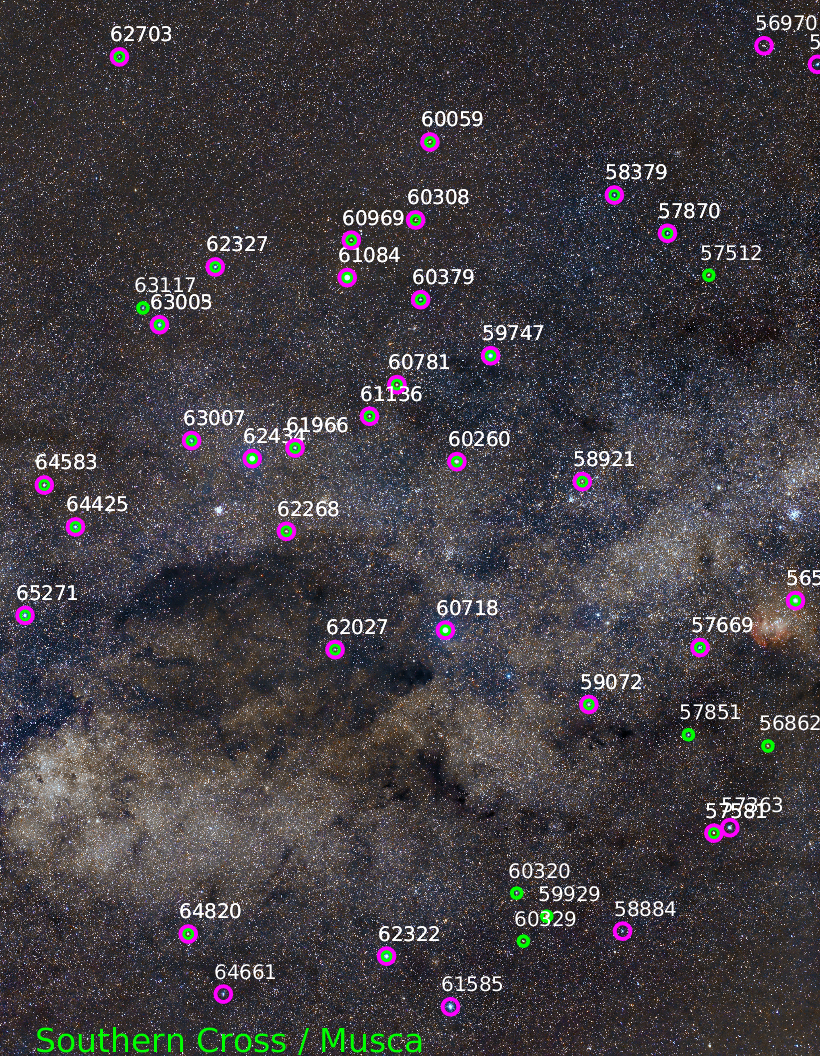}}\\
	\caption{Identification results of ROSIA (green circles) and MPA (magenta circles) on real star images. The constellations in each image are annotated at the bottom part of the image. The identified stars are annotated with their indexes in the Hipparcos catalog.}
\label{fig:real_data_1}
\end{figure*}

\begin{figure*}
    \centering
    \subfloat{\includegraphics[width=0.8\textwidth]{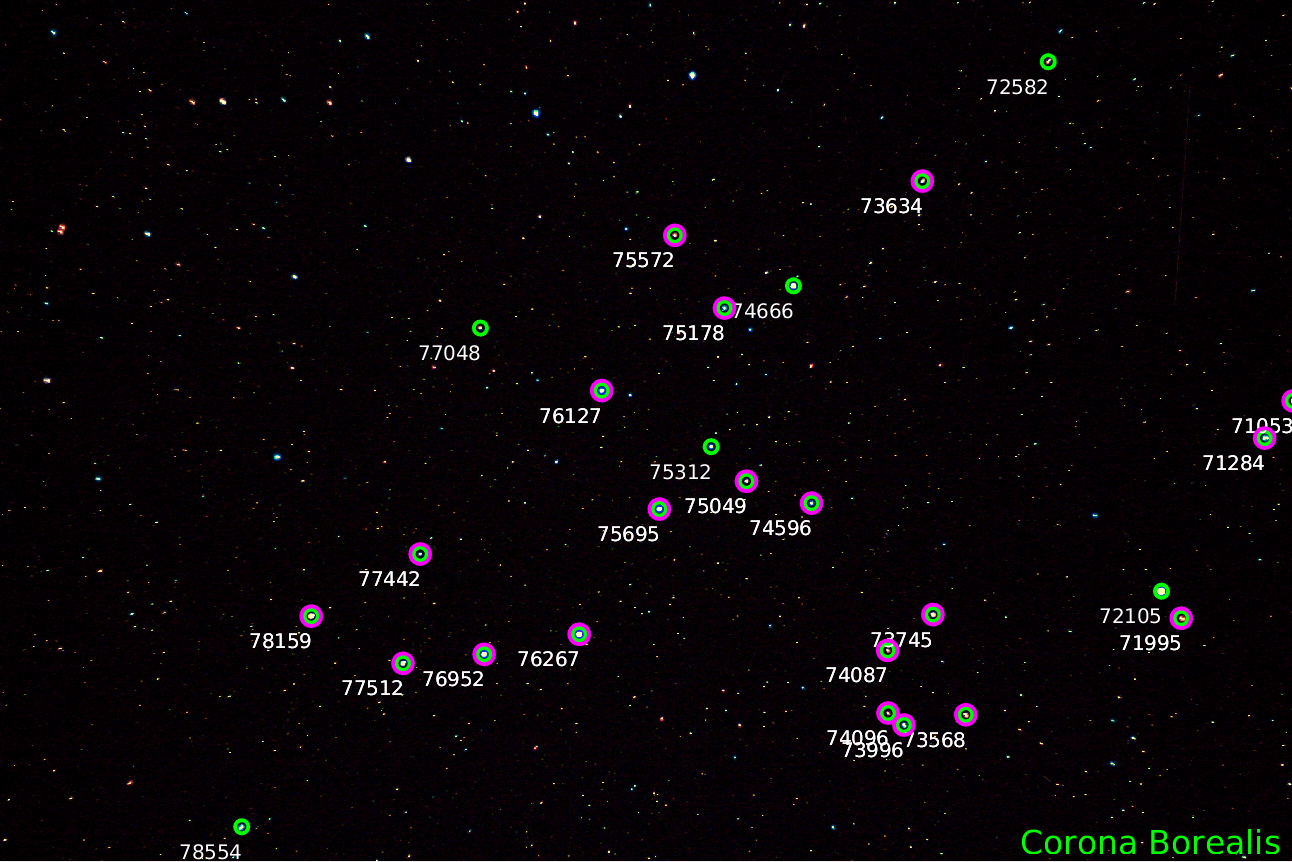}}\\
    \subfloat{\includegraphics[width=0.8\textwidth]{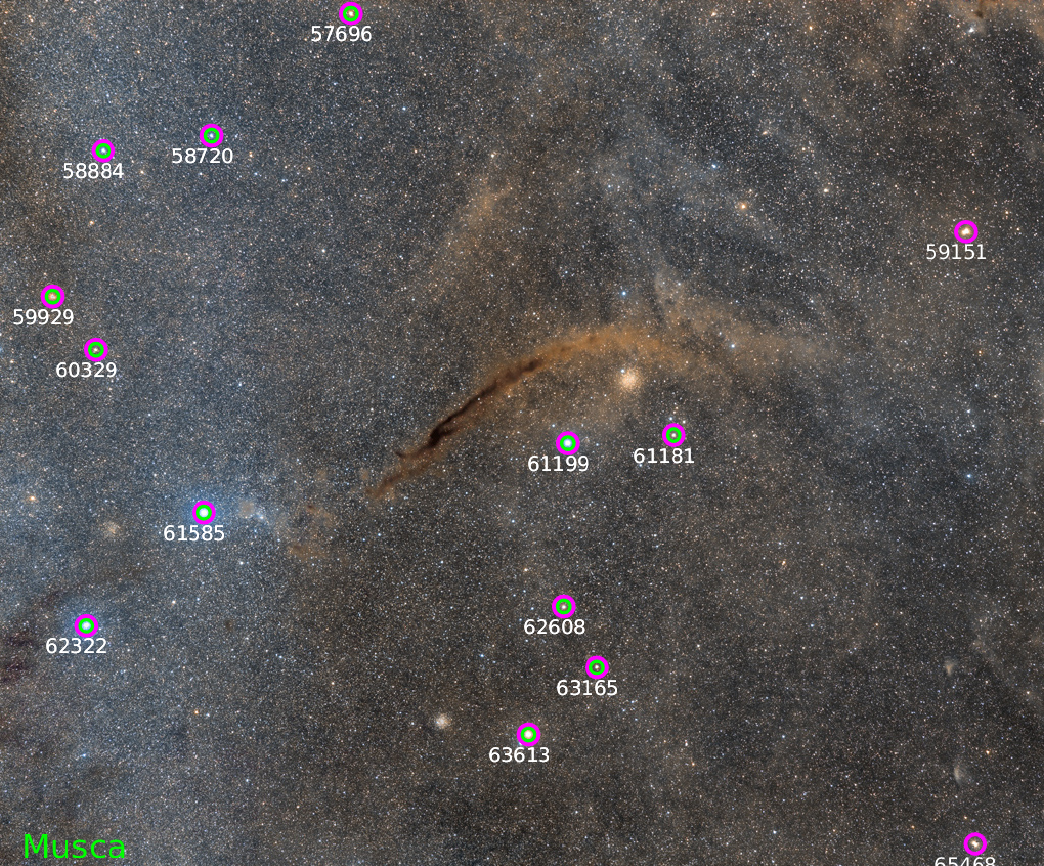}}\\
	\caption{Continuation of Fig.~\ref{fig:real_data_1}}
\end{figure*}

\begin{figure*}
    \centering
    \subfloat{\includegraphics[width=0.8\textwidth]{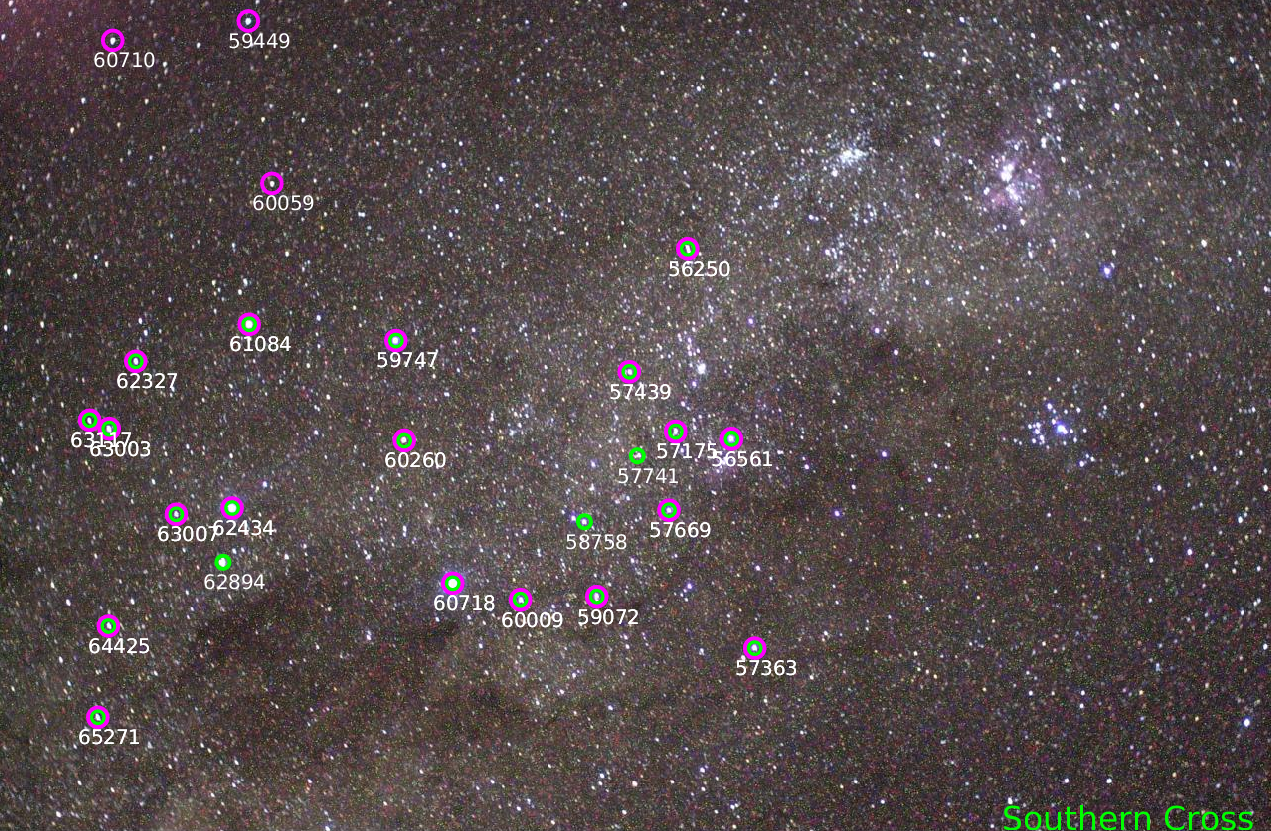}}\\
    \subfloat{\includegraphics[width=0.8\textwidth]{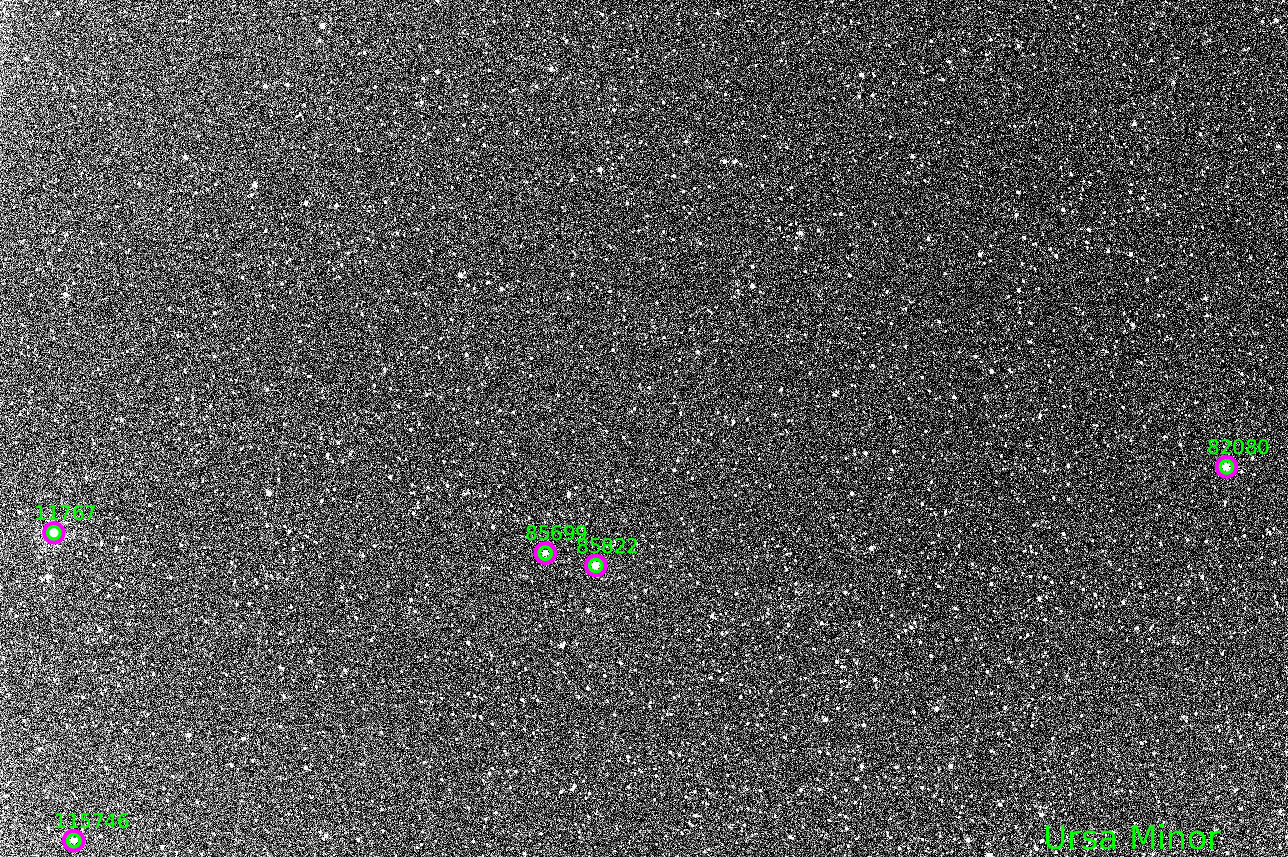}}\\
	\caption{Continuation of Fig.~\ref{fig:real_data_1}}
\end{figure*}

\begin{figure*}
    \centering
    \subfloat{\includegraphics[width=0.8\textwidth]{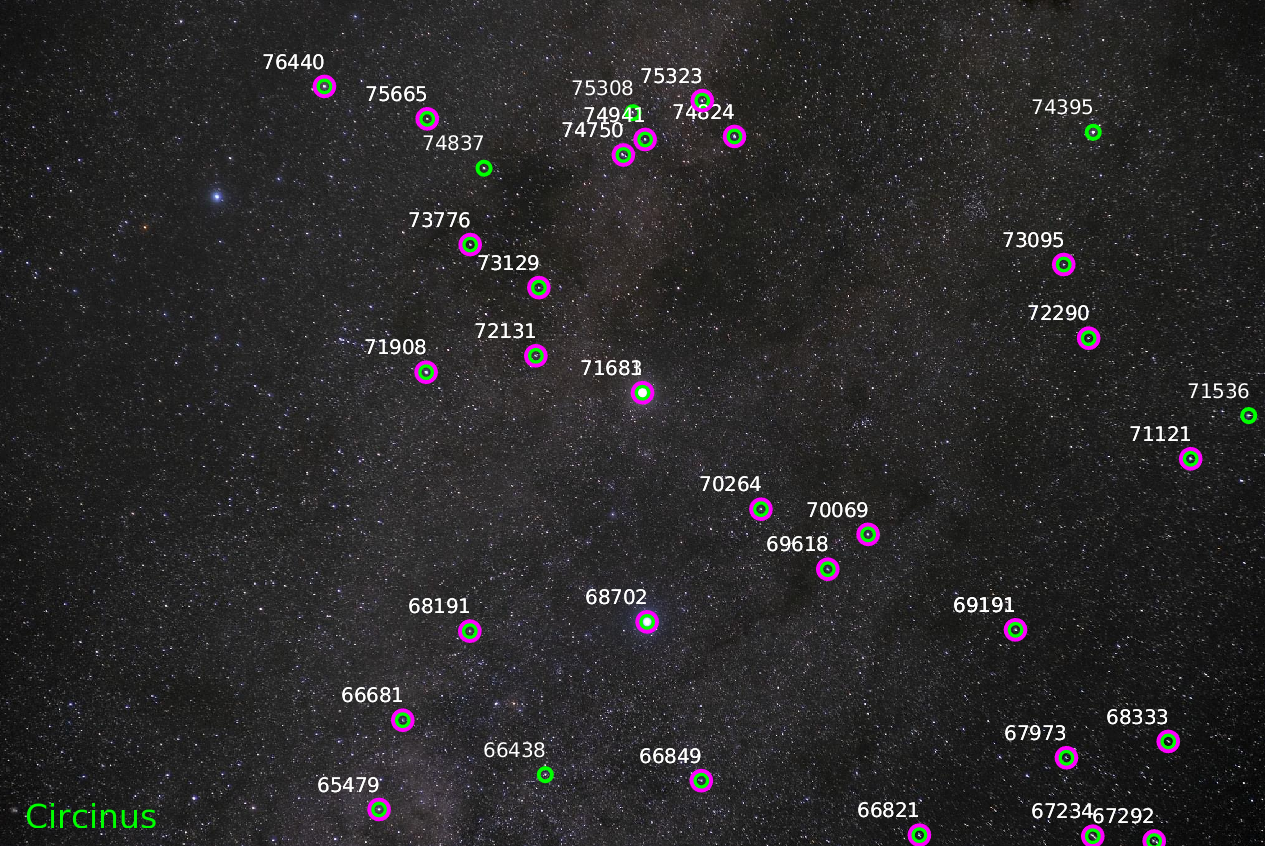}}\\
	\caption{Continuation of Fig.~\ref{fig:real_data_1}}
\end{figure*}

% \end{document}

\end{document}